\documentclass[twocolumn, cleanfoot, 10 pt]{asme2ej}  % Comment this line out if you need a4paper

\usepackage{hyperref}   % to set up hyperlinks
\hypersetup{
	colorlinks=true,
	linkcolor=blue,
	citecolor=blue,
	urlcolor=blue,
}
\usepackage[square,numbers]{natbib}

\usepackage{graphics} % for pdf, bitmapped graphics files
\usepackage{epsfig} % for postscript graphics files
\usepackage{mathptmx} % assumes new font selection scheme installed
\usepackage{times} % assumes new font selection scheme installed
\usepackage{amsmath} % assumes amsmath package installed
\usepackage{amssymb}  % assumes amsmath package installed
\usepackage{kotex}
\usepackage{color}
\usepackage{multirow}
\usepackage{caption}
\usepackage{subcaption}

\bibpunct{[}{]}{,}{n}{}{;}% to get correct punctuation for bibliography
\usepackage[noadjust]{cite}  %for ordering

\graphicspath{{./figures/}}
\DeclareGraphicsExtensions{.pdf,.png,.jpg}

\definecolor{orange}{rgb}{1,0.2,0}
\definecolor{greenish}{rgb}{0,0.6,0}
\newcommand{\cd}[1]{{ {#1}}}
\newcounter{tecounter}
\newenvironment{tightenumerate}
{
	\begin{list}{\arabic{tecounter}\addtocounter{tecounter}{1}.}{%
			\setcounter{tecounter}{1}
			\setlength{\leftmargin}{12pt}
			\setlength{\topsep}{1pt}
			\setlength{\partopsep}{0pt}
			\setlength{\itemsep}{2pt}
			\setlength\labelwidth{7pt}}
		\ignorespaces}
	{\unskip\end{list}
}

\title{\LARGE \bf \cd{Separable Tendon-Driven Robotic Manipulator with a Long, Flexible, Passive  Proximal Section}
}
\author{Christian DeBuys %\thanks{Address all correspondence related to ASME style format and figures to this author.}%\\
    \affiliation{Texas A\&M University\\ Mechanical Engineering\\ College Station, TX, USA\\
        Email: cldebuys@tamu.edu
    }
}
\author{Florin C. Ghesu 
    \affiliation{Siemens Healthineers\\ Digital Technology \& Innovation\\ Princeton, NJ, USA\\
        Email: florin.ghesu@siemens-healthineers.com
    }
}
\author{Jagadeesan Jayender
    \affiliation{Surgical Planning Laboratory\\ Brigham and Women’s Hospital\\ Harvard Medical School, Boston, USA\\
        Email: jayender@bwh.harvard.edu
    }
}
\author{Reza Langari 
    \affiliation{Texas A\&M University\\ Mechanical Engineering\\ College Station, TX, USA\\
        Email: rlangari@tamu.edu
    }
}

\author{Young-Ho Kim \thanks{Corresponding author: Young-Ho Kim} %\\
    \affiliation{Siemens Healthineers\\ Digital Technology \& Innovation\\ Princeton, NJ, USA\\
        Email: young-ho.kim@siemens-healthineers.com
    }
}

\begin{document}
\maketitle
%%%%%%%%%%%%%%%%%%%%%%%%%%%%%%%%%%%%%%%%%%%%%%%%%%%%%%%%%%%%%%%%%%%%%%%%%%%%%%%%
\begin{abstract} % 250 word maximum (currently 163)
%1 broad area of research
\cd{This work tackles} practical issues which arise when using a tendon-driven robotic manipulator\cd{(TDRM)} with a long,\cd{flexible, passive} proximal section in medical applications.\cd{Tendon-driven devices are preferred in medicine for their improved outcomes via minimally invasive procedures, but TDRMs come with unique challenges such as sterilization and reuse, simultaneous control of tendons, hysteresis in the tendon-sheath mechanism, and unmodeled effects of the proximal section shape.}
% \todo{challenges with medical robots: reusable, long tendon sheath mechanism}
%3 proposal by 2-3 sentences : put key terminology (separable TDRM, redundant control, simple linear hysteresis compensation, and proximal shape effect, last sentence is how did we test these things)
% A separable TDRM which overcomes difficulties in actuation and sterilization is introduced.
A separable TDRM which overcomes difficulties in actuation and sterilization is introduced, in which the body containing the electronics is reusable and the remainder is disposable.\cd{An open-loop redundant controller which resolves the redundancy in the kinematics is developed.}
% \cd{An open-loop redundant controller which resolves the redundancy in the kinematics and a physical interpretation of this redundancy are provided.}
\cd{Simple linear hysteresis compensation and re-tension compensation based on the physical properties of the device are proposed.} \cd{The controller and compensation methods are evaluated on a testbed for a straight proximal section, a curved proximal section at various static angles, and a proximal section which dynamically changes angles; and overall, distal tip error was reduced.}
\end{abstract}

\section{Introduction}
Tendon-driven robotic manipulators (TDRM) have been used for various applications, such as remote inspection and maintenance in aerospace, minimally invasive surgery in medicine, and general search and rescue. 
They are preferred in these areas for their ability to maneuver in tight spaces in a compliant and safe manner.
Devices have been developed in the medical domain for cardiac catheterization \citep{stereotaxis20, kim2022automated, reddy_md_view-synchronized_2007}, and bronchoscopy \citep{agrawal_robotic_2020, intuitive_surgical_inc_ion_nodate, johnson__johnson_medtech_monarch_nodate}.
% tendon driven colonosscope???
% Thus, the TSM is an actuation method that has been used in many therapeutic  \citep{daoud_intracardiac_1999, khoshnam_robotics-assisted_2015, khoshnam_robotics-assisted_2017, bai_worldwide_2012, khan_first_2013} and real-time diagnostic (e.g., endoscope \citep{ott_robotic_2011, le_survey_2016, dario_review_2003, phee_locomotion_1997, lee_easyendo_2021}, colonoscope  \citep{chen_development_2006}, and ICE \citep{loschak_algorithms_2017, kim_towards_2021}) manipulators. 
%TEE probe
%2. In medical applications (TDR) --> use ICE (Stereotaxis, siemens), bronchoscopy (Auris, Intuitive), colonoscopy ( different type of catheter (Dupont Table\,1, tendon actuated, examples)
% usecase, How it is difficult to control, common structure of long proximal section [5](Fig2 of [5]), Fig1 (b)(c)
% narrow down like in camarillo 09
Some of these devices are ``flexible-steerable" as classified by Dupont\cd{et al.} \citep{dupont_continuum_2022}, meaning they are comprised of a steerable articulation section and a long, passive proximal section between the articulation section and the actuators.
Such devices are particularly difficult to model and control accurately, and the proximal section is not included in the state-of-the-art kinematic and dynamic models.
Devices with sensors such as Electromagnetic (EM) sensors (e.g., \citep{johnson__johnson_medtech_monarch_nodate}) or Fiber bragg grating (FBG) based sensors (e.g., \citep{intuitive_surgical_inc_ion_nodate, zhang_miniature_2021, cao_closedloop_2022}) can provide a measure of shape and thereby circumvent the unmodeled behavior of the proximal section via closed-loop control. 
However, some devices, such as heart catheters, are one-time or limited-time use, meaning that it would be too costly to implement such sensors and that the unmodeled behavior cannot be compensated as with the aforementioned devices.

\begin{figure*}[t!]
	\centering%\vspace*{-5pt}
	\includegraphics[width=\textwidth]{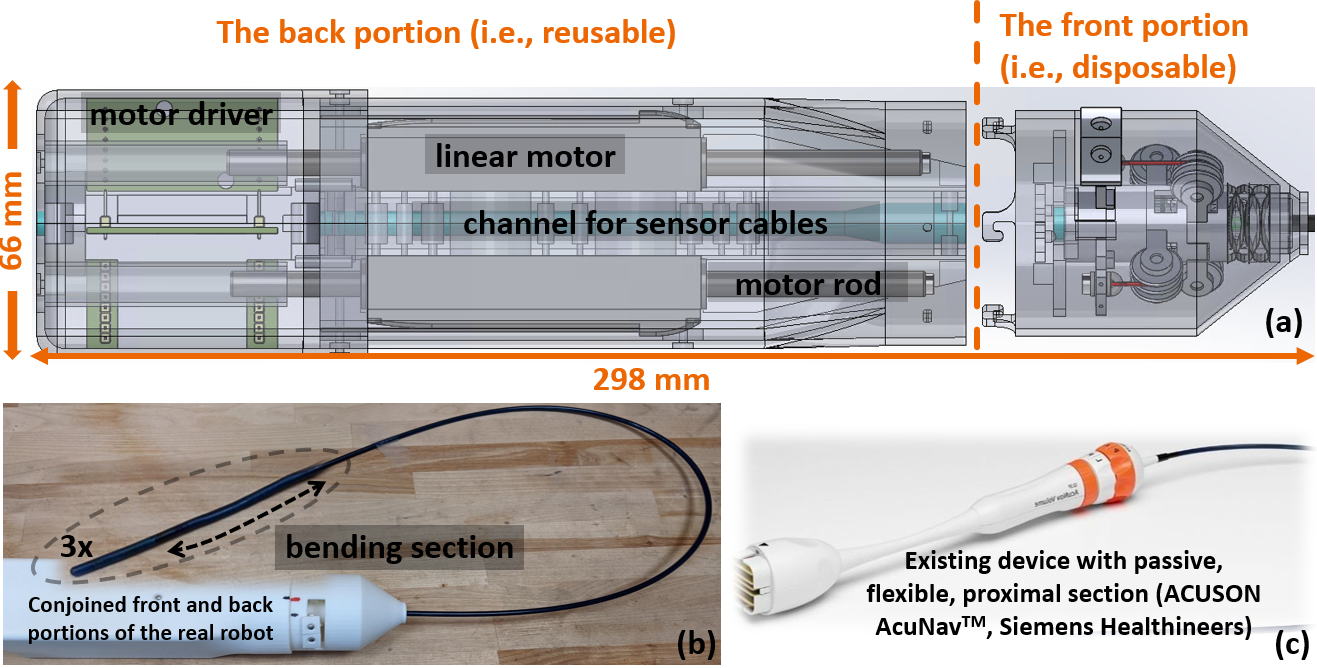}
	\vspace*{0pt}
	%\caption{An overview of the proposed view-to-view robotic manipulation.\label{fig:workflow}}
	\caption{An overview of the catheter robot.}
	\label{fig:robot}
	\vspace*{0pt}
\end{figure*}
%use only fig a and b here, use figure c later in separable design section
%\section{Materials and Methods}

%%%%%%%%%%%%%%%%%%%%%%%%%%%%%%%%%%%%%%%%%%%%%%%%%%%%%%%%

%-why are we using kinematics and CCM (dynamics requires many parameters to be known, determining tension distribution and variable curvature requires more sensor information)

%-most devices in the literature focus on accuracy with less focus/consideration on feasibility (how to say this nicely)

%-why avoid sensors (these devices tend to be disposable, sensors are expensive)

%%%%%%%%%%%%%%%%%%%%%%%%%%%%%%%%%%%%%%%%%%%%%%%%%%%%%%%%%%%%%%
%                       Contribution                         %
%%%%%%%%%%%%%%%%%%%%%%%%%%%%%%%%%%%%%%%%%%%%%%%%%%%%%%%%%%%%%%
%\subsection{Contribution}
%4. So, we build new system; brief decription of novelty 
We introduce a separable TDRM for a practical setting. The separable design tackles the issue of reusability that is common among medical devices, where the part which interacts with the anatomy is disposable and the part containing the actuators is reusable.
% In this paper, we introduce a separable catheter robot with four tendon actuation: one part reusable and another disposable. 
In addition, we utilize a practical model and calibration method for our proposed mechanism so that the four tendons are actuated simultaneously, allowing for precise tip control and mitigating issues with conventional devices, such as dead-zone and hysteresis, with simple linear compensation. 
We consider an open-loop controller since many available devices \citep{kim2022automated,lee21hysteresis} are used without position-tracking sensors at the tip due to costs and single use. 
We analyze the effect of the shape of the passive proximal section for different compensation types and offer insight on how this behavior might be accounted for in open and closed-loop systems.

\vspace*{-0pt}

%%%%%%%%%%%%%%%%%%%%%%%%%%%%%%%%%%%%%%%%%%%%%%%%%%%%%%%%%%%%%%
%                       Related Works                        %
%%%%%%%%%%%%%%%%%%%%%%%%%%%%%%%%%%%%%%%%%%%%%%%%%%%%%%%%%%%%%%
\section{Related Works}  %% +20 papers  ~25th (Tuesday)
% 5. surgical/device robot in TDM devices--> how this works in practical mechanical settings.--> why we design separable structures (LOOK AT NONLINEAR HYSTERESIS PAPER RELATED WORKS)
The tendon-sheath mechanism facilitates maneuverability and compliance in highly constrained anatomies by allowing for a long, thin, flexible, proximal tail-like structure.
The TDRM is an actuation method that has been used in many therapeutic  \citep{daoud_intracardiac_1999, khoshnam_robotics-assisted_2015, khoshnam_robotics-assisted_2017, bai_worldwide_2012, khan_first_2013, hasanzadeh_efficient_2014} and real-time diagnostic (e.g., endoscope \citep{ott_robotic_2011, le_survey_2016, dario_review_2003, phee_locomotion_1997, lee_easyendo_2021}, and Intracardiac echocardiography (ICE) \citep{loschak_algorithms_2017, kim_towards_2021, zhongyu2021}) manipulators. 
% Thus, the TSM is an actuation method that has been used in many therapeutic  \citep{daoud_intracardiac_1999, khoshnam_robotics-assisted_2015, khoshnam_robotics-assisted_2017, bai_worldwide_2012, khan_first_2013} and real-time diagnostic (e.g., endoscope \citep{ott_robotic_2011, le_survey_2016, dario_review_2003, phee_locomotion_1997, lee_easyendo_2021}, colonoscope  \citep{chen_development_2006}, and ICE \citep{loschak_algorithms_2017, kim_towards_2021}) manipulators. 
The increased compliance of tendon-driven manipulators is not without drawbacks. Much work has gone into accurately modeling the kinematics; accounting for phenomena such as hysteresis, deadzone, and slack; and overcoming the redundancy in the control input.
With respect to forward kinematics, many publications have presented a lumped-parameter approach with the constant curvature model (CCM) \citep{hasanzadeh_efficient_2014, kim_towards_2021, rone_mechanics_2014, sitler_modular_2022, camarillo_mechanics_2008, xu_investigation_2008}. Reviews for modeling tendon-driven continuum manipulators have been given \citep{rao_how_2021, webster_design_2010}, but the shape of the passive proximal section--which can be very long for devices such as catheters--is not included in these kinematic models. 
Shape sensing is possible and has been reviewed \citep{shi_shape_2017, amanzadeh_recent_2018}, but the sensors necessary (e.g. FBG or EM) would be prohibitively expensive for a disposable device or disposable portion of a device.
Model-based shape estimation which considers friction and external force has been developed \citep{li_modelbased_2020}, but the model does not include a long, passive proximal section.
We take first steps to characterize the effect of the proximal section on bending angle error of the articulation/bending section.

%7. 
Hysteresis has been addressed for TDRMs via compensation methods \citep{do_hysteresis_2014, xu_motion_2016, wang_active_2020, kato_extended_2014} and modeling approaches \citep{li_modelbased_2020, do_hysteresis_2014, zglimbea_identification_nodate, hassani_structural_2013}. Since the modeling approaches involve many hyper-parameters, which themselves require a complicated identification process, Lee et al. \citep{lee21hysteresis} provided a simplified hysteresis model which included deadzone and backlash. 
Kim et al. \citep{kim_shape-adaptive_2021} proposed a practical shape-adaptive hysteresis compensation based on dead-zone detection using motor current, where compensation is adjusted based on arbitrary shape change of the proximal shaft. 
We implement a simple hysteresis compensation which does not require hyper-parameters and a redundant control input which removes deadzone.
% do not add to the hysteresis literature and instead

%8. 
Redundant control strategies for tendon-driven devices have been implemented for cable driven parallel robots \citep{platform}, dexterous robot hands \citep{abdallah_decoupled_2013}, and continuum manipulators \citep{camarillo}. 
Fang et al. \citep{platform} handled platform and cable dynamics, provided a redundant control input with tension and actuator constraints to prevent slack, and solved an optimization real-time to determine the control input. However, due to their different application, their tendons were free floating and thus did not have tendon-sheath friction. 
Abdallah et al. \citep{abdallah_decoupled_2013} handled multi-joint finger and cable dynamics with an optimization similar to \citep{platform}, but with the addition of tendon-sheath friction. 
Camarillo\cd{et al.} \citep{camarillo} gave a redundant control scheme for quasi-static motion of a catheter with two articulation sections that decoupled the inverse kinematics. Their optimization was similar to \citep{platform, abdallah_decoupled_2013} except the solution allowed for slack tendons to be present (as long as they contribute no force). 
We propose a redundant control input which prevents slack tendons as in \citep{platform, abdallah_decoupled_2013} but does not require an optimization and which yields a physical interpretation of the redundancy. 
Unlike \citep{platform, abdallah_decoupled_2013}, the state transition matrix ($\textbf{C}$ in this paper) is not dependent on the configuration of the manipulator, resulting in a single solution to the inverse kinematics for a given desired configuration.
We also present simplified kinematics for an incompressible articulation section.
% Hysteresis models (DH LEe papers --> non-linear hysteresiss model-based one (Bocwen) -->in real setting problem; long proximal section structure/require identification in real-time [what to talk about here, we arent doing anything spectacular with the hysteresis input, so how much do we want to mention?]

% Dead zone present in all commercial catheters, talk about prior works and how redundant control input minimizes the deadzone naturally (talk about it in methods-discussion) [again, how much do we want to talk about this, the controller naturally addresses it but we do not explicitly show the difference with and without the redundant control]

%%%%%%%%%%%%%%%%%%%%%%%%%%%%%%%%%%%%%%%%%%%%%%%%%%%%%%%%%%%%%%
%                           Methods                          %
%%%%%%%%%%%%%%%%%%%%%%%%%%%%%%%%%%%%%%%%%%%%%%%%%%%%%%%%%%%%%%
\section{Materials \& Methods} 

%Summary of subsections

% The following subsections will cover the design of the robot, the derivation of the redundant control scheme and determination of its parameters, the simple hysteresis and re-tension compensations, and the experimental setup. 
The following subsections will cover the design of the robot (Section\,\ref{sec:design}), the derivation of the redundant control scheme and determination of its parameters (Section\,\ref{sec:redundant}), and the simple hysteresis and re-tension compensations (Section\,\ref{sec:hysteresis}). 

%\subsection{Design of Separable Tendon-Driven Continuum Manipulator} \label{sec:design}
\subsection{Design of Separable Tendon-Driven Robotic Manipulator}\label{sec:design}

\cd{The existing manual device is shown in Fig.\,\ref{fig:robot}(c). This disposable catheter is actuated by hand by a physician; each knob connects to two opposing tendons and controls motion in one bending plane. Due to the structure of the handle, each knob can only pull one of its tendons at a time, depending on the direction of rotation, and a deadzone or area of slack can develop as the tendons stretch. The proposed device tackles the issue of reusability by introducing a separable mechanism and of deadzone by actuating all tendons. In short, the proposed device differs in that it is actuated, all four tendons can be manipulated independently, and it is separable to avoid disposing of the expensive electronics.}
% Figure 1(c) is the type of manual device that our prototype roboticizes. The proposed device differs in that it is actuated, all four tendons can be manipulated independently, and it is separable to avoid disposing of the expensive electronics.

%\textcolor{blue}{use both names, back and reusable, in this section}

%Fig.\,\ref{fig:robot} gives an overview of the design.
The proposed robotic system shown in Fig.\,\ref{fig:robot} consists of two parts: the {\em back portion} or {\em reusable portion} contains four Faulhaber linear motors (LM 1483-080-11-C) with motor drivers mounted to a plastic core. 
% The plastic core contains a channel along its central axis for an ultrasound (US) cable and its connector, which are shown in blue in Fig.\,\ref{fig:robot}(a). 
The plastic core contains a channel along its central axis for an ultrasound (US) cable or for any other sensor cables, which is shown in blue in Fig.\,\ref{fig:robot}(a). 
%The rods of the linear motors have magnets fastened to their ends to facilitate attachment of the tendons during clipping. The linear motors actuate the tendons by moving the rods forward and back. 
\begin{figure}[t!]
	\centering%\vspace*{-5pt}
	\includegraphics[width=\columnwidth]{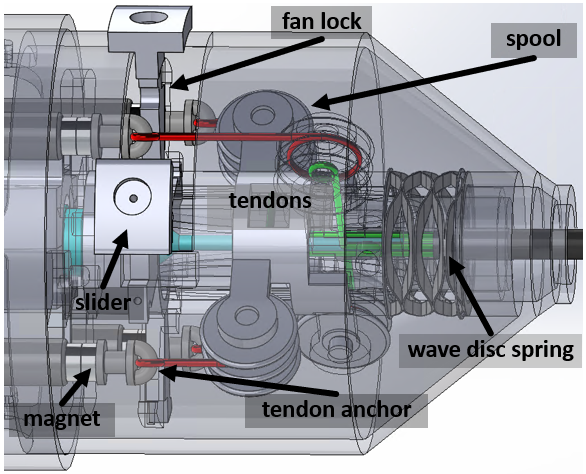}
	\vspace*{0pt}
	%\caption{An overview of the proposed view-to-view robotic manipulation.\label{fig:workflow}}
	\caption{Disposable portion of the robot.}
	\label{fig:front}
	\vspace*{0pt}
\end{figure}
\begin{figure}[t!]
	\centering%\vspace*{-5pt}
	\includegraphics[width=0.40\columnwidth]{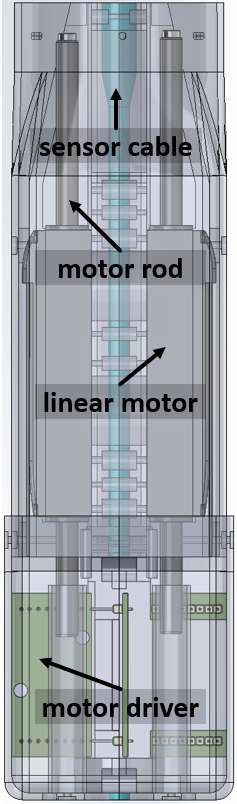}
	\vspace*{0pt}
	%\caption{An overview of the proposed view-to-view robotic manipulation.\label{fig:workflow}}
	\caption{Reusable portion of the robot.}
	\label{fig:back}
	\vspace*{0pt}
\end{figure}
\cd{The {\em front portion} or {\em disposable portion} includes the tendons, the passive proximal section, and the articulation section, all of which are taken from an existing catheter as shown in Fig.\,\ref{fig:robot}(c).}
%\yh{Any update here? Otherwise, finish the sentence!}\todo{sheath is made of ptfe, tendons are made of vectran, check existing papers for materials}
% The {\em front portion} or {\em disposable portion} contains all of the tendons and includes the catheter's passive, proximal section and articulation section.
% \cd{The tendons, proximal section, and distal articulation section are taken from an existing catheter as shown in Fig.\,\ref{fig:robot}.}
The interface from one tendon in the disposable portion shown in Fig.\,\ref{fig:front} to its respective linear motor is as follows: 1) the tendon (green) from the catheter is bent 90 degrees around a low-friction roller and is wrapped around and fastened to the small radius of the spool, 2) a separate tendon (red) is wrapped around and fastened to the large radius of the spool and then attached to the tendon anchor, and 3) the tendon anchor is held in place by the fan lock until the catheter is clipped together and the motor rod has connected to the tendon anchor via magnetic force. 
\cd{The \textit{fan lock} is specially designed for this device and is so named because the ``blades" of the fan are used to lock the tendon anchors in place until it is opened.}
% \yh{Is it specifically only 3:1? How to describe as general? Reasoning! What is this spool doing?} \yh{Where is wave-disc spring?} \textcolor{blue}{just say it increases pulling force based on ratio} \textcolor{blue}{mention wave disc spring which is for clamping}
The spools serve to increase the pulling force of the motors with a pulley ratio of $R$:$r$, where $R$ is the larger radius and $r$ is the smaller radius; and in our case the ratio is 3:1. The wave-disc spring serves to push the front portion against a locking mechanism consisting of plastic hooks in the back portion after the system has been clamped.

%{\em ``The front portion"} contains a ``fan-lock'' and a stacked wave-disc spring mounted on a plastic core. 
%the \hl{catheter itself}, four plastic "tendon anchors" with magnets attached, plastic spools to increase the pulling force (3:1) of the motors, plastic "sliders", and a plastic cover. The tendons from the catheter (shown in green) are attached to the smaller radius of the spools, the tendon anchors are attached to the larger radius of the spools by a separate tendon (shown in red), and the tendon anchors are held in place by the fan-lock until they are manually released after clipping.
The proposed design has several advantages:
\begin{tightenumerate}
\item[1)] Direct actuation of the tendons via linear motors facilitates a more transparent control input and gives a reliable measure of tendon tension via motor current.
% the next sentence is a complete sentence by itself and causes problems
This prevents the need for additional tension sensors and reduces the complexity of hysteresis phenomena compared to \citep{lee21hysteresis}, simplifying the control problem.
\item[2)] The reusable portion contains no tendons, meaning that the issue of tendon wear present in many devices is avoided. 
The disposable portion (Fig.\,\ref{fig:front}) can be unclipped and reclipped indefinitely, since the fan lock mechanism can clamp the tendons in place when the disposable portion is not attached and since the interface between the motor rods and the tendon anchors is magnetic (it does not involve a single use fastener).
\item[3)] The clipping system has a closing mechanism similar to that of a child-proof medicine bottle: the front portion is pushed into the body until it makes contact, pushed slightly farther and twisted, then released. 
After releasing, the front is locked, the motors move their rods toward the front until their magnets make contact with the magnets on the tendon anchors, the sliders are pushed away from each other to manually open the fan-lock, and the motors are free to pull on the tendons.
\end{tightenumerate}

\subsection{Derivation of redundant control}\label{sec:redundant}
\subsubsection{Moment balance equation}

% \begin{figure}%[t!]
% 	\centering%\vspace*{-5pt}
% 	\includegraphics[width= \columnwidth]{figures/catheter-both.PNG}
% 	\vspace*{-0pt}
% 	%\caption{An overview of the proposed view-to-view robotic manipulation.\label{fig:workflow}}
% 	\caption{Conceptual side-view and cross-section}
% 	\label{fig:catheter-both}
% 	\vspace*{-0pt}
% \end{figure}

\begin{figure}%[t!]
	\centering%\vspace*{-5pt}
	\includegraphics[width= \columnwidth]{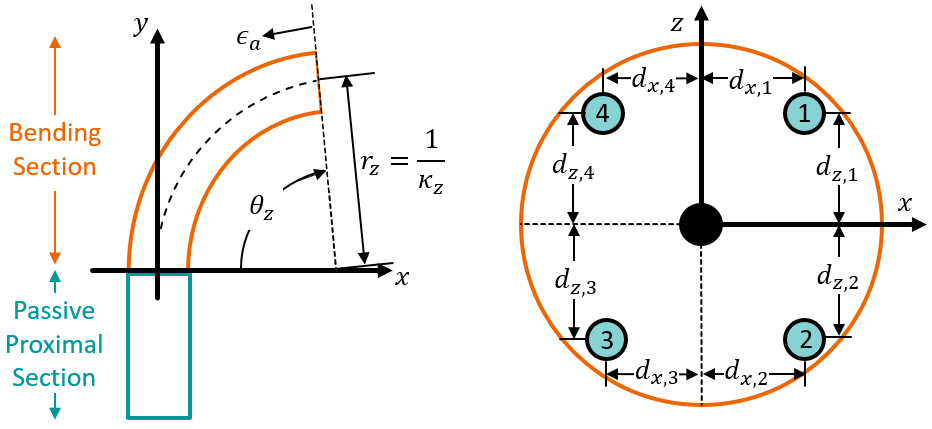}
	\vspace*{-0pt}
	%\caption{An overview of the proposed view-to-view robotic manipulation.\label{fig:workflow}}
	\caption{Conceptual side-view and cross-section. $\theta_z$, $\kappa_z$, and $r_z$ are the bending angle, curvature, and radius of curvature about the z-axis (in the xy-plane). $(d_{xi},d_{zi})$ are the coordinates of tendon \textit{i} relative to the central axis (y-axis) of the robot.}
	\label{fig:catheter-both}
	\vspace*{-0pt}
\end{figure}

We borrow nomenclature from \citep{camarillo} and define the robot's configuration for a compressible bending section as $\cd{\textbf{q}} = [\kappa_{x}, \kappa_{z}, \epsilon_{a}]^T$, where curvature about the x-axis $\kappa_{x}$ [1/mm] corresponds to bending in the yz-plane, curvature about the z-axis $\kappa_{z}$ [1/mm] corresponds to bending in the xy-plane, and axial strain $\epsilon_a$ [unitless] corresponds to compression along the y-axis.  %\yh{What is the curvature? and unit?,  the magnitude of the derivative of the unit tangent vector? , one divided by the radius of curvature?}
Just as in \citep{camarillo}, curvature is chosen rather than bending angle for the configuration variables\cd{in $\textbf{q}$} to obtain linear kinematic and static equations. 
First, we write the moment balance equation for a compressible bending section and $n$ tendons:
\begin{equation}
\begin{bmatrix} K_b~~~0~~~~0 \\ ~0~~~K_b~~~0 \\ ~~0~~~~0~~~K_a \end{bmatrix} \begin{bmatrix} \kappa_x \\ \kappa_z \\ \epsilon_a \end{bmatrix} = \begin{bmatrix}  ~~d_{z1}~~~~d_{z2}~ \cdots ~~~ d_{zn} \\ -d_{x1}~ -d_{x2}~\cdots -d_{xn} \\ ~~~~~1 ~~~~~~1~ \cdots ~~~~~1 \end{bmatrix} \begin{bmatrix} T_1 \\ T_2 \\ \vdots \\ T_n \end{bmatrix},
\label{eq_moment}
\end{equation}
where the matrix $\textbf{K}$ is a stiffness matrix containing $K_b$ [N$\cdot$mm$^2$], which is bending stiffness with respect to curvature $\kappa$, and $K_a$ [N], which is axial stiffness with respect to strain $\epsilon_a$.
The matrix $\textbf{D}$ results from the cross product of the tendon locations $(d_{xi},d_{zi})$ [mm] with the tendon tensions $T_i$ [N] where $i \in 1, 2, \dots, n$. A conceptual side-view of the xy-plane and cross-section of the xz-plane are shown in Fig.\,\ref{fig:catheter-both}.
% where $K_b$ [N$\cdot$mm$^2$] is bending stiffness with respect to curvature $\kappa$, $K_a$ [N] is axial stiffness with respect to strain $\epsilon_a$, $\textbf{D}$ is a matrix resulting from the cross product of the tendon locations [mm] with the tendon tensions $T$ [N]. A conceptual side-view of the xy-plane and cross-section for two tendons are shown in Fig.\,\ref{fig:catheter-both}.

%Note a few notation differences in Eq.\,\eqref{eq_moment} compared to [Camarillo], namely that our numbering starts from 1 instead of 0, and that the bending axes are x and y instead of x and z (resulting in some sign changes).
If the bending section undergoes negligible compression, we can deviate from Eq.\,\eqref{eq_moment} and reduce the robot's configuration to $\textbf{q} = [\kappa_{x}, \kappa_{z}]^T$. The corresponding moment balance for an incompressible bending section is: %\yh{When can we ignore compression? Please give more clear exaplanation. I am curious}
% \yh{q is vector -->change to bold, Check all the things vectors or scalar,,,, D, K as well Definition too!}
% We define the robot's configuration $q = [\kappa_{x}, \kappa_{z}]^T$, where curvature is chosen rather than bending angle to obtain linear kinematic and static equations as in \citep{camarillo}, and axial compression is considered negligible. 
%If we assume the bending section is incompressible, then we can adjust the moment balance equation.
\begin{equation}
\begin{bmatrix} K_b~~0 \\ ~0~~K_b \end{bmatrix} \begin{bmatrix} \kappa_x \\ \kappa_z \end{bmatrix} = \begin{bmatrix}  ~~d_{z1}~~~~d_{z2}~ \cdots ~~~ d_{zn} \\ -d_{x1}~ -d_{x2}~\cdots -d_{xn} \end{bmatrix} \begin{bmatrix} T_1 \\ T_2 \\ \vdots \\ T_n \end{bmatrix}.
\label{eq_moment_incompressible}
\end{equation}
We can write Eq.\,\eqref{eq_moment} or Eq.\,\eqref{eq_moment_incompressible} in matrix form:
\begin{equation}
\textbf{K} \textbf{q} = \textbf{D} \tau .
\label{eq_moment_matrix}
\end{equation}
The relation in Eq.\,\eqref{eq_moment_matrix} describes the static equilibrium, where the tendon tensions must balance against each other and the spring-like forces inherent to the bending of the catheter.

Let $dim(\tau) = m$ and $dim(\textbf{q}) = n$. When $m > n$, the system is redundant since there are more inputs $m$ than outputs $n$. 
This means, for a given desired configuration $\textbf{q}_d$, there are infinitely many solutions $\tau$ to Eq.\,\eqref{eq_moment_matrix}.
% \yh{What is the difference between Camarillo's approach for FK with slack? He mentioned "in situations where slack may occur, we must ensure positive tendon tension by more advanced optimization techniques." Then  he moved to FK with slack. I would like to know what is the difference between this. And, Do we have Inverse Kinematics? As he described?}
% \cd{For slack, yes, they have a section dedicated to getting a solution with slack tendons. They give a constraint in the previous section for positive tension but say "However, these simple methods do not account for slack and therefore should be used with caution. In situations where slack may occur, we must ensure positive tendon tension (6) by more advanced optimization techniques"
% So the problem they are alluding to, is that even with $\tau \geq 0$ constraint, you can still run into slack tendons.
% Their solution is to introduce a method which can get feasible solutions even with slack tendons.
% Our solution is to use the structure of the redundant control input and introduce mu, which is chosen to get a certain amount of pre-tension.}
It is possible to solve this redundancy by minimizing the control effort via real-time optimization, but this will not\cd{necessarily} solve the issue of slack tendons. We could include the constraint $\tau \geq 0$ in the optimization, but we can develop a control scheme which resolves the redundancy and avoids real-time optimization.
% We could include $\tau \geq 0$ as a constraint in an optimization which solves for minimum $\tau$ as in \citep{platform, abdallah_decoupled_2013}but this problem can be solved more explicitly in a way which leverages the physical meaning of the redundancy in the system to prevent tendon slack.
% \yh{TODO: Move or Locate here?}
% \textcolor{blue}{Prior works have included $\tau \geq 0$ as a constraint in an optimization which solves for minimum $\tau$ in cable-driven platform \citep{platform} and dexterous hand applications \citep{abdallah_decoupled_2013}, and Camarillo et al. \citep{camarillo} developed an algorithm which optimizes $\tau$ while allowing slack tendons (as long as they contribute no force) for a continuum manipulator. We differ from these works in that we leverage the structure of the constant curvature kinematics to get an analytical solution and avoid real-time optimization altogether. We also differ from \citep{camarillo} since our method does not permit slack tendons.}
%\yh{Are we going to talk about both compressible bending section andnoncompressible? What does it mean in  our paper?} 
%We could include $\tau \geq 0$ as a constraint in an optimization which solves for minimum $\tau$ as in \citep{platform, abdallah_decoupled_2013} or , but this problem can be solved more explicitly in a way which leverages the physical meaning of the redundancy in the system to prevent tendon slack.
Derivation of the control scheme is handled for both the \textit{compressible} and \textit{incompressible} cases\cd{for completeness, although the results in this work will only use the incompressible case}.
%\yh{Is this our novel idea? or any reference do we have? Then, add or emphasize this is different with reference}
%\cd{paragraph before this comment was adjusted to address this question}

\subsubsection{Compressible bending section (n=3, m=4)}

%Suppose we have four tendons equidistant from each other and from the central axis of the beam as in Fig.\,\ref{fig:cross}. 
Suppose we have four tendons located at 90 degree increments around the central axis of the beam. 
% \begin{figure}%[t!]
% 	\centering%\vspace*{-5pt}
% 	\includegraphics[scale= 0.6]{figures/catheter-cross-section.PNG}
% 	\vspace*{-0pt}
% 	%\caption{An overview of the proposed view-to-view robotic manipulation.\label{fig:workflow}}
% 	\caption{Conceptual cross-section of the manipulator}
% 	\label{fig:cross}
% 	\vspace*{-0pt}
% \end{figure}
If the bending section is compressible, then we are using Eq.\,\eqref{eq_moment} and have three outputs. With one more input than output, we have one dimension of redundancy. Thus, the family of solutions for a given desired output can be parameterized along one vector, namely the vector which spans the nullspace of the map from the input to the output. We reorganize Eq.\,\eqref{eq_moment} and define $\textbf{C} = \textbf{K}^{-1} \textbf{D}$, which is the map from the input $\tau$ to the output $q$:
\begin{equation}
\textbf{q} = \textbf{C} \tau.
\label{eq_C}
\end{equation}
For a fully controllable system, $rank(\textbf{C}) = n$, and the locations of the tendons in this system render it fully controllable (even if tendons can only pull, i.e., even if $\tau \geq 0$). If we label the last tendon as redundant, we can partition $\textbf{C}$ into a full rank portion $\textbf{B}$, which is square, and a redundant portion $\textbf{h}$, which is a column vector, as in \citep{platform}: 
% \yh{Don't we need to address reference? finger paper idea?} \textcolor{blue}{add note about how this is like platform paper}
\begin{equation}
\textbf{q} = [\textbf{B}~~\textbf{h}] \tau.
\label{eq_Bh}
\end{equation}
Given a desired output $\textbf{q}_d$, the control input $\tau$ is:
\begin{equation}
\tau = \begin{bmatrix} -\textbf{B}^{-1} \textbf{h} \\ 1 \end{bmatrix} \mu + \begin{bmatrix} \textbf{B}^{-1} \textbf{q}_d \\ 0 \end{bmatrix},
\label{eq_redundant}
\end{equation}
where the desired output with a scalable parameter $\mu$ has no affect on the output. In other words, our family of solutions is parameterized by $\mu$ along the direction of the null space vector $\textbf{n} = [-\textbf{B}^{-1} \textbf{h} ~~ 1]^T$.  The variable $\mu$ does not affect the output because $\textbf{n}$ is in the null space of  $\textbf{C}$:
\begin{equation}
\textbf{C} \textbf{n} = [\textbf{B}~~\textbf{h}] \begin{bmatrix} -\textbf{B}^{-1} \textbf{h} \\ 1 \end{bmatrix} = \textbf{0}.
\label{eq_null_verification}
\end{equation}
The problems of redundancy and preventing tendon slack are solved with Eq.\,\eqref{eq_redundant} and an appropriate choice of $\mu$. An obvious choice for $\mu$ is the one that minimizes the control effort \citep{kirk_optimal_2004, liberzon_calculus_2012} as follows:
%\yh{what is fraction? 1/2} \textcolor{blue}{give a reference for this cost function, maybe a textbook}
\begin{equation}
\begin{aligned}
\min_{\mu} \quad & \frac{1}{2}\sum_{i=1}^{n} T_i ^2 ,\\
\textrm{s.t.} \quad &T_i\geq0,    \\
\end{aligned}
\label{eq_opt}
\end{equation}
where the constraints on $T_i$ (no slack) and the bottom row of Eq.\,\eqref{eq_redundant} imply that $\mu \geq 0$. The choice of control input in Eq.\,\eqref{eq_redundant} allows us to directly calculate $\mu_{min}$ instead of solving this optimization computationally. In the unconstrained case, the minimum is found as follows:
\begin{equation}
\sum_{i=1}^{n} T_i \frac{\partial T_i}{\partial \mu} = 0.
\label{eq_opt_min}
\end{equation}
% \textcolor{blue}{explain that this optimization would normally be solved real-time for other people, but with this setup we can find 1 mu which works and never need to optimize again}
If the solution to Eq.\,\eqref{eq_opt_min} does not violate the constraints in Eq.\,\eqref{eq_opt}, then that minimum is sufficient. %\yh{Sounds a bit oo informal} 
If it does violate those constraints, then we simply pick the smallest $\mu$ which satisfies all of the constraints. 
%In this case, $\mu$ can be found by checking $T_i = 0$ and choosing the largest $\mu \geq 0$ computed from those equations.
%\textcolor{blue}{Partition \textbf{B} into rows $\beta_i$ and give the minimum equations explicitly.}
To highlight, let the rows of $\textbf{B}^{-1}$ be denoted by $\beta_i$ where $i \in 1, 2, 3$ such that $\textbf{B}^{-1} = [\beta_1 ~\beta_2 ~\beta_3]^T$. Then we can write the constraints from Eq.\,\eqref{eq_opt} for tension in each tendon:
% \begin{equation}
% T_i = -\beta_i \textbf{h} \mu + \beta_i \textbf{q}_d \geq 0 ~~~ \textrm{for} ~~~ i = 1,2,3
%     \label{eq_opt_min_comp_explicit}
% \end{equation}
\begin{equation}
\begin{split}
T_i = -\beta_i \textbf{h} \mu + \beta_i \textbf{q}_d \geq 0 ~~~ \textrm{for} ~~~ i \in 1,2,3 ~~, \\
\mu \geq 0, ~~~~~~~~~~~~~~~~~~~~~~~~~~~
\label{eq_opt_min_comp_explicit}
\end{split}
\end{equation}
where we can see that every term in the equations except for $\mu$ is determined by the structure of the robot ($\beta_i$ and $\textbf{h}$) and the desired configuration ($\textbf{q}_d$).
The strength of this controller is that, as long as the structure of the robot in $\textbf{C}$ does not vary explicitly with configuration or time, we can choose a constant value for $\mu$ and achieve a control input for any desired state $\textbf{q}_d$ without real-time optimization.
% where the largest $\mu \geq 0$ yielded from setting the equations in Eq.\,\eqref{eq_opt_min_comp_explicit} to 0 is the one which ensures all tendons are in tension.

% \yh{Are we talking about specific our robot in method section?}
% \textcolor{blue}{reword in our caseandmake more general}
We elaborate on the previous statement and note the limitations of the analysis so far. In our case, the choice of redundant tendon and thus the redundant column of $\textbf{C}$ was arbitrary; we could choose any tendon and get the same result. This is because the locations of the tendons relative to the central axis do not depend on the system configuration, which means $\textbf{B}$ is independent from the input. In other robots \citep{platform}, it is possible for $\textbf{B}$ to become ill-conditioned with changing configuration, in which case an algorithm is required to ensure an appropriate tendon is chosen.\cd{The final step is to derive the kinematics, but first the case for an incompressible articulation section--which is more relevant in this work--must be derived.}

\subsubsection{Incompressible bending section (n=2, m=4)}

% \yh{For our catheter?? This seems one example?} \textcolor{blue}{reword to not mention our device specifically}
In this case, the compression of the bending section is negligible compared to the magnitude of bending. Therefore, we can use Eq.\,\eqref{eq_moment_incompressible} and will have two outputs. 
This results in two dimensions of redundancy and a family of solutions which can be parameterized by two scalars, $\mu_1$ and $\mu_2$, along the directions of two vectors, $\textbf{n}_1$ and $\textbf{n}_2$, which span the null space of $\textbf{C}$. We arbitrarily label the last two tendons as redundant and partition $\textbf{C}$ into a full rank portion $\textbf{B}$, which is square, and two redundant portions $\textbf{h}_1$ and $\textbf{h}_2$, which are both column vectors. For a given desired output $\textbf{q}_d$, the control input $\tau$ as follows:
\begin{equation}
\tau = \begin{bmatrix} -\textbf{B}^{-1} \textbf{h}_1 \\ 1 \\ 0 \end{bmatrix} \mu _1 + \begin{bmatrix} -\textbf{B}^{-1} \textbf{h}_2 \\ 0 \\ 1 \end{bmatrix} \mu _2 + \begin{bmatrix} \textbf{B}^{-1} \textbf{q}_d \\ 0 \\ 0 \end{bmatrix},
\label{eq_redundant2}
\end{equation}
where $\mu_1$ and $\mu_2$ are the desired output with two scalable parameters. As before, adjusting $\mu_1$ or $\mu_2$ has no affect on the output since $\textbf{n}_1 = [-\textbf{B}^{-1} \textbf{h}_1 ~~ 1 ~~ 0]^T$ and $\textbf{n}_2 = [-\textbf{B}^{-1} \textbf{h}_2 ~~ 0 ~~ 1]^T$ lie in the null space of $\textbf{C}$:
\begin{equation}
\begin{aligned}
\textbf{C} \textbf{n}_1 = [\textbf{B}~~\textbf{h}_1~~\textbf{h}_2] \begin{bmatrix} -\textbf{B}^{-1} \textbf{h}_1 \\ 1 \\ 0 \end{bmatrix} = \textbf{0}, \\
\textbf{C} \textbf{n}_2 = [\textbf{B}~~\textbf{h}_1~~\textbf{h}_2] \begin{bmatrix} -\textbf{B}^{-1} \textbf{h}_2 \\ 0 \\ 1 \end{bmatrix} = \textbf{0}.
\end{aligned}
\label{eq_null_verification2}
\end{equation}
If the tendons are located equidistant from each other about the central axis, the assumption of incompressibility decouples the bending caused by one pair of tendons (1 and 3) from the bending caused by the other pair of tendons (2 and 4). In other words, $\textbf{n}_1 = [1~0~1~0]^T$ and $\textbf{n}_2 = [0~1~0~1]^T$, which means $\mu_1$ and $\mu_2$ do not appear in the same tension equation. Thus, the minimum $\mu_1$ and $\mu_2$ can be determined independently, in a procedure similar to the compressible case via Eq.\,\eqref{eq_opt} and Eq.\,\eqref{eq_opt_min}. 
%\textcolor{blue}{Partition \textbf{B} into rows $\beta_i$ and give the minimum equations explicitly.}
If we denote the rows of $\textbf{B}$ by $\beta_i$ such that $\textbf{B} = [\beta_1 ~ \beta_2]^T$, we can write as follows: 
\begin{equation}
\begin{split}
T_i = -\beta_i \textbf{h} \mu_i + \beta_i \textbf{q}_d \geq 0 ~~~ \textrm{for} ~~~ i \in 1,2 ~~, \\
\mu_1 \geq 0, ~~~~~~~~~~~~~~~~~~~~~~~ \\
\mu_2 \geq 0, ~~~~~~~~~~~~~~~~~~~~~~~
    \label{eq_opt_min_incomp}
\end{split}
\end{equation}
where $\mu_1$ and $\mu_2$ are the only unknowns and can be chosen independently. 
Again we see that we need only pick $\mu_1$ and $\mu_2$ to obtain feasible tensions without real-time optimization.
% \begin{equation}
% T_i = -\beta_i \textbf{h} \mu + \beta_i \textbf{q}_d ~~~ \textrm{for} ~~~ i = 1,2,3 \\
%     \label{eq_opt_min_incomp2}
% \end{equation}
% \yh{exact number here?} \textcolor{blue}{remove exact numbers here}
For a minimum allowable tension of $0$ (i.e., no slack condition), $\mu_1 = \mu_2 = 0$, but the constraint $\tau \geq 0$ is generally not strict enough to ensure that no slack occurs. 
%We found $0.25$ N, and thus $\mu_1 = \mu_2 = 0.25$, to be a low pre-tension which prevents slack tendons. 
Therefore, small constants should be found experimentally for $\mu_1$ and $\mu_2$ to ensure a low pre-tension which prevents slack tendons.
Why this is referred to as "pre-tension" will be addressed in the discussion.
%\yh{Do we need this section here? or moving to discussion?}
%\cd{Moving that paragraph to discussion}
% \subsubsection{Physical interpretation of the redundancy}
% Mathematical redundancy in real systems often points to a physical phenomenon. 
% % If we consider Fig.\,\ref{fig:cross} and the incompressible case, the redundancy represents the co-contraction or simultaneous pulling of opposing tendons, such as tendons 1 and 3 or tendons 2 and 4, just as opposing tendons are pulled in muscle contractions. 
% If we assume the four tendons are at 90 degree increments and incomressibility, the redundancy represents the co-contraction or simultaneous pulling of opposing tendons, such as tendons 1 and 3 or tendons 2 and 4, just as opposing tendons are pulled in muscle contractions.
% The magnitude of this co-contraction is determined by $\mu_1$ and $\mu_2$. In theory, $\mu_1$ and $\mu_2$ could be any positive value without affecting the configuration of the manipulator. In practice, large values of $\mu_1$ or $\mu_2$ could cause compression of the "incompressible" bending section, buckling of the passive proximal section, and changes in the stiffness estimates. Thus, it is best to use small values of $\mu_1$ and $\mu_2$ as a method for preventing slack tendons in that any $\mu_1$ or $\mu_2$ value above zero effectively enforces a pre-tension in the corresponding tendons equal to the magnitude of $\mu_1$ or $\mu_2$. 
With $\mu_1$ and $\mu_2$ discussed, we move on to the kinematics and summarize the procedure for determining the control input.

\subsubsection{Kinematics}
\begin{figure}%[t!]
	\centering%\vspace*{-5pt}
	\includegraphics[width = 0.6\columnwidth]{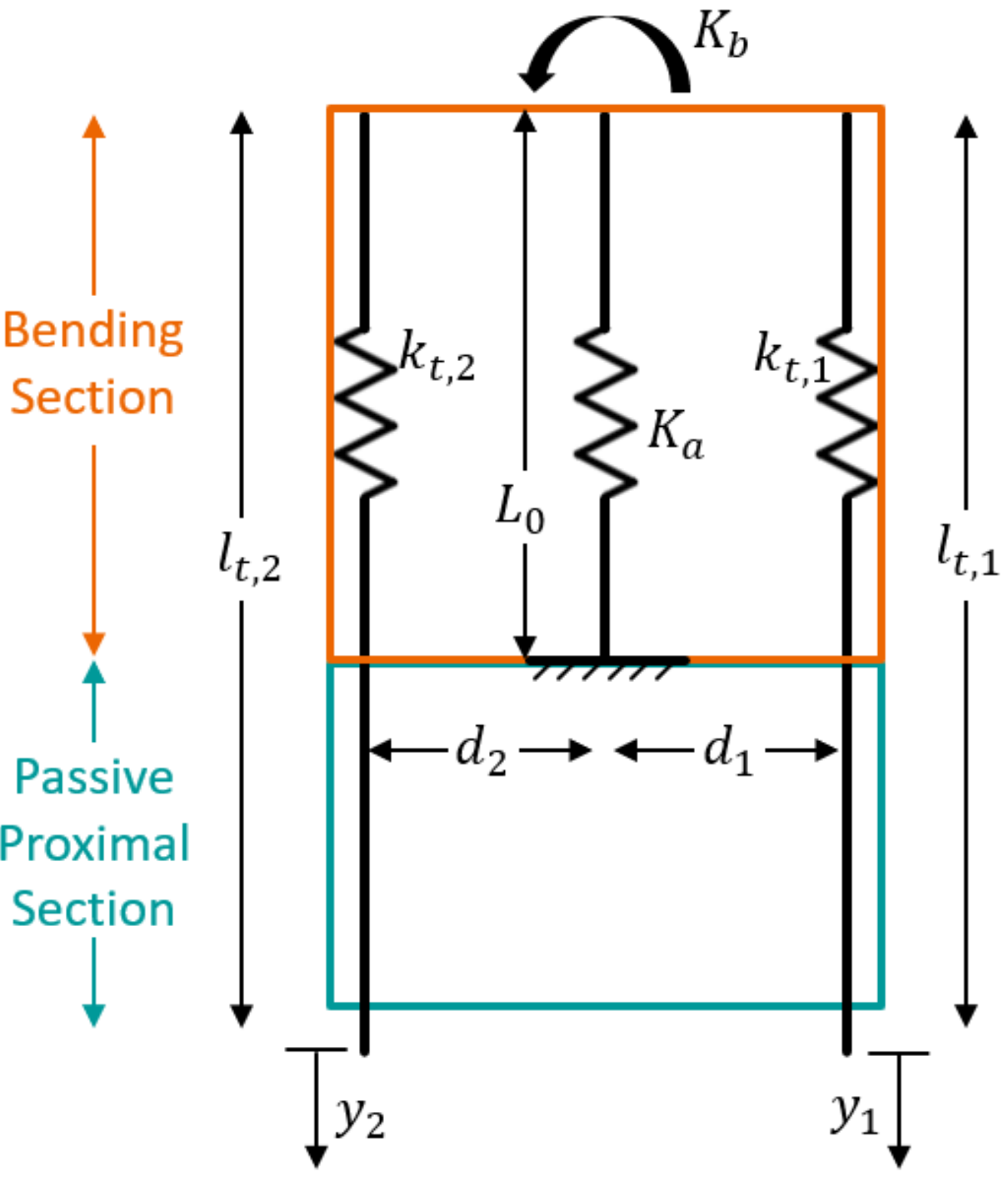}
	\vspace*{-0pt}
	%\caption{An overview of the proposed view-to-view robotic manipulation.\label{fig:workflow}}
	\caption{Planar spring model for two tendons: $L_0$, $K_a$, and $K_b$ are the length, axial stiffness, and bending stiffness of the articulation section; $l_{t,i}$, $k_{t,i}$, $d_i$, and $y_i$ are the undeformed length, stiffness, distance from the central axis, and displacement of tendon \textit{i}.}
	\label{fig:planar-spring}
	\vspace*{-0pt}
\end{figure}

%\yh{Reference camillio, and be nice to explain how to get equation below Y=g$\tau$ What is y D}
%\cd{addressed}
The preceding discussion yielded the force input (tendon tensions) for the system. However, it is often preferred to command actuator positions (tendon displacements). To convert the input from forces $\tau$ to positions $\textbf{y}$, we consider the conservation of strain equation \citep{camarillo}: 
% \textcolor{blue}{add strain equation}
\begin{equation}
\textbf{y} = \textbf{D}^T \textbf{L}_0 \textbf{q} + \textbf{L}_t \textbf{K}_t^{-1} \tau,
\label{eq_strain}
\end{equation}
where $\textbf{L}_0$, $\textbf{L}_t$, and $\textbf{K}_t$ are diagonal matrices containing the undeformed bending section length $L_0$, the unstretched tendon lengths $l_{t,i}$, and the length-normalized tendon stiffnesses $k_{t,i}$, respectively. For instance, tendon 1 would have unstretched length $l_{t,1}$ located in row 1 and column 1 of $\textbf{L}_t$. 
The moment balance in Eq.\,\eqref{eq_moment_matrix} along with Eq.\,\eqref{eq_strain} define a planar spring model in which the moments and forces exerted by the tendons balance against the inherent bending and axial stiffness of the articulation section. An example for a two-tendon planar spring model is given in Fig.\,\ref{fig:planar-spring}.
Using Eq.\,\eqref{eq_moment_matrix} we can finally obtain a direct relation between $\textbf{y}$ and $\tau$:
\begin{equation}
\textbf{y} = (\textbf{D}^T \textbf{L}_0 \textbf{K}^{-1} \textbf{D} + \textbf{L}_t \textbf{K}_t^{-1}) \tau.
\label{eq_strain_tau}
\end{equation}

If the actuators pulled the tendons directly, Eq.\,\eqref{eq_strain_tau} would be sufficient. Since our system uses pulleys to increase the effective pulling force of the actuators, we have additional steps. The force felt by the motors $\tau_m$ is given by: 
% The displacement of the motors $\textbf{y}_m$, which depends on the displacement of the catheter tendons $\textbf{y}$ and the deflection of the additional tendons due to the motor force $\tau_m$, is given in Eq.\,\eqref{eq_y_spools}.
\begin{equation}
\tau_m = \textbf{R}^{-1} \tau,
\label{eq_tau_spools}
\end{equation}
where $\textbf{R}$ is a diagonal matrix containing the pulley ratio.

Then, the displacement of the motors $\textbf{y}_m$, which depends on the displacement of the catheter tendons $\textbf{y}$ and the deflection of the additional tendons due to the motor force $\tau_m$, is as follows:
\begin{equation}
\textbf{y}_m = \textbf{R} \textbf{y} + \textbf{L}_m \textbf{K}_t^{-1} \tau_m,
\label{eq_y_spools}
\end{equation}
where $\textbf{L}_m$ is a diagonal matrix containing the undeformed lengths of the additional tendons $l_{m,i}$, which are shown in red in Fig.\,\ref{fig:front}. 
Substituting Eq.\,\eqref{eq_strain_tau} into Eq.\,\eqref{eq_y_spools} yields:
\begin{equation}
\textbf{y}_m = \textbf{R} (\textbf{D}^T \textbf{L}_0 \textbf{K}^{-1} \textbf{D} + \textbf{L}_t \textbf{K}_t^{-1}) \tau + \textbf{L}_m \textbf{K}_t^{-1} \tau_m,
\label{eq_y_spools_intermediate}
\end{equation}
and replacing $\tau$ with $\tau_m$ using Eq.\,\eqref{eq_tau_spools}, we can rewrite Eq.\,\eqref{eq_y_spools} in terms of the displacement of the actuators $\textbf{y}_m$ and forces felt by the actuators $\tau_m$: 
%\textcolor{blue}{give intermediate steps} %ratio*(ratio*G + Lm*Km.inverse())
\begin{equation}
\textbf{y}_m = (\textbf{R}^2 (\textbf{D}^T \textbf{L}_0 \textbf{K}^{-1} \textbf{D} + \textbf{L}_t \textbf{K}_t^{-1}) + \textbf{L}_m \textbf{K}_t^{-1}) \tau_m.
\label{eq_strain_tau_spools}
\end{equation}
% $\textbf{L}_m$ is a diagonal matrix containing the undeformed lengths of the additional tendons $l_{m,i}$, which are shown in red in Fig.\,\ref{fig:front}.
Now that we have Eq.\,\eqref{eq_strain_tau_spools}, the process of determining the control input can be summarized.

\subsubsection{Redundant control summary}

The scalars $\mu_1$ and $\mu_2$ are found experimentally and are chosen such that tendons remain in tension. Since these values can effect parameter estimates, it is best to leave them constant once they are chosen. Given $\mu_1$ and $\mu_2$, the tendon tensions in $\tau$ which correspond to the desired configuration $\textbf{q}_d$ are computed using Eq.\,\eqref{eq_redundant} or Eq.\,\eqref{eq_redundant2}, depending on whether the manipulator is compressible. The tensions in $\tau$ are converted into tendon displacements in $y$ using Eq.\,\eqref{eq_strain_tau}. This method gives a single vector $\textbf{y}$ for any desired configuration $\textbf{q}_d$, resolving the redundancy while preventing slack tendons. The redundant control input is considered the \textit{baseline} input for the remainder of this work.
%With the baseline control input established, we move on to the compensation methods. %\yh{I like this paragraph!}

% \yh{TODO: Move or Locate here?}
Prior works have included $\tau \geq 0$ as a constraint in an optimization which solves for minimum $\tau$ in cable-driven platform \citep{platform} and dexterous hand applications \citep{abdallah_decoupled_2013}. Camarillo\cd{et al.} \citep{camarillo} developed an algorithm which optimizes $\tau$ while allowing slack tendons (as long as they contribute no force) for a continuum manipulator. We differ from these works in that we leverage the structure of the constant curvature kinematics to get an analytical solution and avoid real-time optimization altogether. We also differ from \citep{camarillo} since our method does not permit slack tendons.
With the baseline control input established, we move on to the compensation methods. %\yh{I like this paragraph!}
%Additional compensation for hysteresis and passive proximal section angle are described in the next section.

% The goal is to compute the tendon tensions which achieve the desired output, convert those tensions into tendon displacements, command the tendon displacements, and ensure that the tendons remain in tension. 
% To implement the redundant control, it is necessary to choose $\mu_1$ and $\mu_2$ based on the minimum allowable tension. For a minimum allowable tension of $0$ (i.e., no slack condition), $\mu_1 = \mu_2 = 0$. We found $0.25$ N, and thus $\mu_1 = \mu_2 = 0.25$, to be a low pre-tension which prevents slack tendons. The algorithm is as follows:
% \begin{enumerate}
%   \item Compute the tendon tensions which correspond to the desired configuration $q_d$ using Eq.\,\eqref{eq_redundant} or Eq.\,\eqref{eq_redundant2}.
%   \item Convert those tendon tensions into tendon displacements using Eq.\,\eqref{eq_strain_tau}
% \end{enumerate}
% - describe calculating exact minimum and adding safety factor to tension (Do we need tension plots?)

%\subsection{Parameter identification}

%\subsection{Simple hysteresis and re-tension compensation}

% \subsection{Practical Hysteresis Compensation methods}
% \cd{is this a suggested name change for the following subsection?}

\subsection{Practical Compensation methods}\label{sec:hysteresis}

\subsubsection{Preliminaries: Time-shift and DC Offset}

Before introducing the compensation methods, we must mention the two phenomena which they will compensate. 
\begin{figure}[h!]
     \centering
     \begin{subfigure}{0.86\columnwidth}
         \centering
         \includegraphics[width=\textwidth]{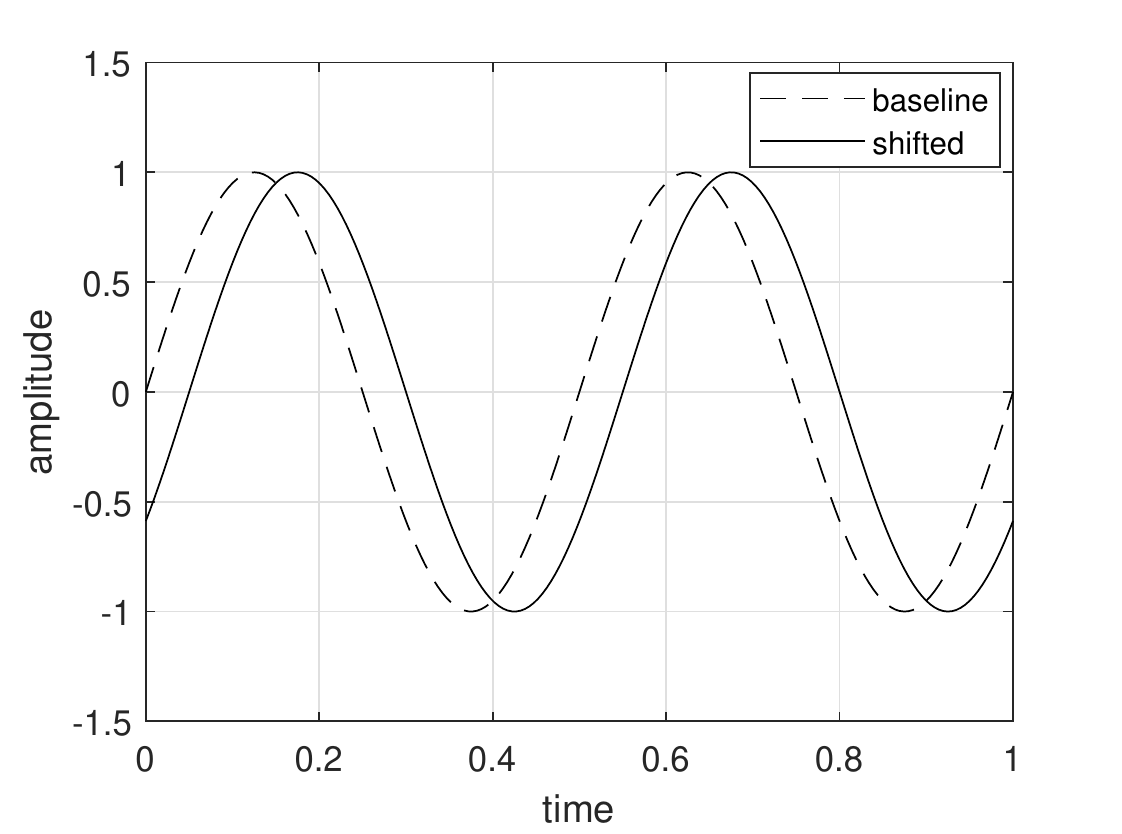}
         \caption{Idealized time-shift due to hysteresis}
         \label{fig:example-shift}
     \end{subfigure}
     %\hfill
     \begin{subfigure}{0.86\columnwidth}
         \centering
         \includegraphics[width=\textwidth]{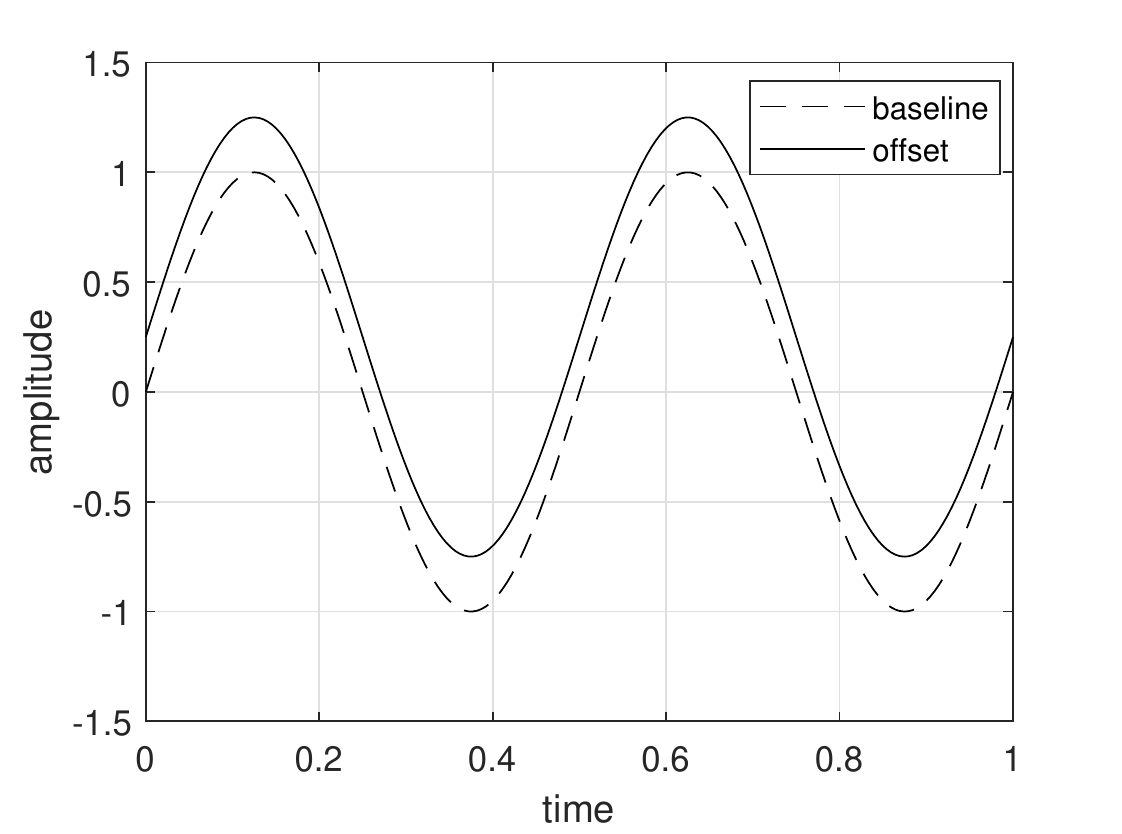}
         \caption{Idealized DC offset due to proximal section angle}
         \label{fig:example-offset}
     \end{subfigure}
        \caption{Phenomena in need of compensation.}
        \label{fig:example-phenomena}
\end{figure}
\cd{\textit{Hysteresis} is defined as the phenomenon in which the value of a physical property lags behind changes in the effect causing it.}
In the presence of unmodeled hysteresis, an output will tend to lag an input in an open-loop system each time the direction of the input changes, due to \cd{friction induced} backlash from the hysteresis. 
For a periodic input, this lag manifests as an obvious phase lag or \textit{time-shift} as idealized in Fig.\,\ref{fig:example-shift}. 
% For a periodic input, this lag manifests as an obvious phase lag or \textit{time-shift} as idealized in Fig.\,\ref{fig:example-shift} and in Eq.\,\eqref{eq_def_hyst}.
We hypothesize that hysteresis compensation can reduce this phenomena and thereby reduce error. 
% \yh{Why do you propose simple hysteresis models?} \textcolor{blue}{just say this is what hysteresis comp is for, dont mention ours specifically}
% \begin{equation}
% \begin{split}
% x_{in} = A sin(\omega t), ~~~~~~~~~\\
% x_{out} = A sin(\omega (t - t_s)).
% \end{split}
% \label{eq_def_hyst}
% \end{equation}
% \textcolor{blue}{why does this happen, give example of catheter in anatomy, motivate}
Bending of the passive proximal section will naturally occur during operation of such flexible-steerable devices, for instance in traversing blood vessels to reach the heart in catheter applications, and this behavior is typically unmodeled.
We found that introducing an angle in the proximal section in one direction causes the articulation section to bend in the other direction, as visible when comparing Fig.\,\ref{fig:sub-setup1} with \ref{fig:sub-setup2}. We guess that this is caused by the increased stretching and thus increased tension in the outer tendon(s) and decreased tension in the inner tendon(s). This change in tensions causes an offset of the unbent configuration of the articulation section, which results in an offset of the output. For a sinusoidal input in the same plane as the proximal section deformation, the offset in the output would manifest as a \textit{DC offset} of the sinusoid as shown in Fig.\,\ref{fig:example-offset}. 
% \yh{Bring motivational story from introduction, just reminder it, and why it exist in practice?}

We further predict that these two sources of error are, for the most part, independent. For a sinusoidal input $x_{in}$ with amplitude $A$ and frequency $\omega$, we would expect the output $x_{out}$ to be shifted in time due to hysteresis by some amount $t_s$ and offset due to the passive proximal section angle by some amount $A_s$, as follows:
\begin{equation}
\begin{split}
x_{in} = A sin(\omega t), ~~~~~~~~~\\
x_{out} = A sin(\omega (t - t_s)) + A_s.
\end{split}
\label{eq_sin}
\end{equation}

\subsubsection{Simple hysteresis compensation for time-shift}
\begin{figure}%[h]
	\centering%\vspace*{-5pt}
	\includegraphics[scale= 0.35]{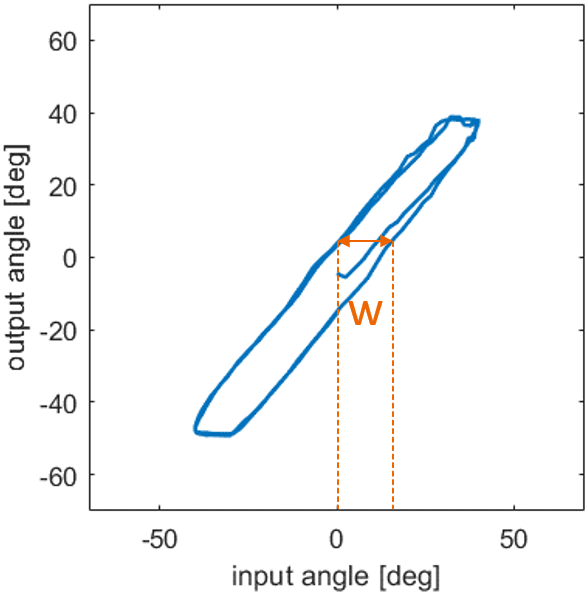}
	\vspace*{-0pt}
	%\caption{An overview of the proposed view-to-view robotic manipulation.\label{fig:workflow}}
	\caption{Hysteresis curve with width \textit{w}.}
	\label{fig:hyst-width}
	\vspace*{-0pt}
\end{figure}

% \yh{due to proposed structure (full actuation) , DEFINITIION of Hysteresis curve with width \textit{w}} \yh{If you add w/2 then does it still satisfy the constraints in controls? Address this!} \textcolor{blue}{explain that we can use simple hysteresis comp because our mechanism and control scheme create a simple hysteresis curve as seen in figure 7. we could handle more complicated models such as buoc when but in this paper we focus on a simple approach which can be used without extra sensors and without high fidelity parameter identification}
The design of the robot and the control scheme result in a hysteresis curve, shown in Fig.\,\ref{fig:hyst-width}, which is very linear and has approximately no deadzone. The simplicity of this hysteresis curve allows us to implement a simplified hysteresis compensation on top of the control scheme which does not rely on extra sensors or tuning a large number of parameters; however, those methods are compatible with the control scheme and could still be implemented in future works. 

The simple hysteresis compensation consists of a constant value added to the desired input depending on the direction of motion. This constant value is based on the width $w$ of the hysteresis curve, also shown in Fig.\,\ref{fig:hyst-width}, which can be determined from the input-output map of a continuum manipulator and is device dependent. 
% A hysteresis curve is shown in Fig.\,\ref{fig:hyst-width}. 
The input and output are given as angles in degrees since bending angle is easier to conceptualize than curvature, and the compensation equation is also given as an angle:
\begin{equation}
\theta_{d,h} = \theta_d + \frac{w}{2}.
\label{eq_hyst}
\end{equation}

This desired angle with hysteresis compensation $\theta_{d,h}$ is converted into a desired curvature using the usual equation for constant curvature in the incompressible case:
\begin{equation}
\kappa_{d,h} = \frac{\theta_{d,h}}{L_0}.
\label{eq_th2k}
\end{equation}

This additional compensation involves an adjustment of the input configuration in order to achieve the configuration which is truly desired, and thus it completely compatible with the previously described control scheme.
% There are more complex methods of compensating hysteresis, but the goal is to keep the compensation simple so that it does not obscure any sources of error and so that it does not require additional sensors or intensive hysteresis parameter identification.

\subsubsection{Re-tension compensation for DC offset}
% \yh{We haven't talked about proximal shape effect yet. What is the proximal shape and how does it effect?} \cd{addressed, since we have added it into the introduction}

% \textcolor{blue}{link this to DC offset via title and first sentence}
The re-tension compensation aims to remove error introduced by the passive proximal section angle, which manifests as a DC offset for a sinusoidal input. This angle can come about naturally in medical applications, such as in the tortuous path of a catheter from the incision point to the heart.
In the baseline case, the robot is initialized, and the passive proximal section angle is introduced with no movement of the actuators. 
That is, the motors are powered and holding their positions under the assumption that the proximal section is straight. 
This is not unreasonable, as current kinematic and dynamic representations of continuum robots with bending sections make the assumption that the passive proximal section is straight. 
In the case with re-tension compensation, the robot is given approximately 3-5 seconds to re-tension the tendons to the desired pre-tension (0.25 N) after the passive proximal section angle has been introduced but before data for the sinusoidal input is collected. 
Motor current is filtered using a $3^{rd}$ order Butterworth filter to remove high frequency noise, and this filtered motor current is used along with the force constant from the manufacturer to get a measure of motor force and thus of tendon tension.
It is anticipated that re-tensioning the tendons will remove some of the error from the passive proximal section angle.

% \textcolor{red}{Do we need to talk about the low level PID controller which tracks the minimum tension set-point?} \yh{give the steps how does it work in practice} \textcolor{blue}{explain that motor current was measured directly and converted to force using constant from manufacturer. motor current was filtered using butterworth, describe that the setpoint is desired tension}

%%%%%%%%%%%%%%%%%%%%%%%%%%%%%%%%%%%%%%%%%%%%%%%%%%%%%%%%%%%%%%%%%
%                           Experiment                          %
%%%%%%%%%%%%%%%%%%%%%%%%%%%%%%%%%%%%%%%%%%%%%%%%%%%%%%%%%%%%%%%%%
\section{Experiments and Results}
% \textcolor{blue}{explain that we only use incompressible case but we present both in case someone wants to use this control scheme for a compressible device}

% \cd{It was expected that error would increase for increasing proximal section angle due to the DC offset introduced and that hysteresis would increase error at all angles, and simple compensation techniques are proposed to address these error sources. Hysteresis compensation is expected to reduce time delay, while re-tension compensation should reduce DC offset, which will likely be the primary source of increasing error. The re-tension compensation was also tested for dynamic changes in the proximal section and was expected to outperform the baseline case.
% % Bending angle error increased for increasing proximal section angle for all testing conditions with an average error reduction of 41.48\% for re-tension, 4.28\% for hysteresis, and 52.35\% for re-tension + hysteresis compensation relative to the baseline case. 
% % Two major sources of error in tracking the bending angle were identified: time delay from hysteresis and DC offset from the proximal section angle. 
% }
% \todo{give roadmap to section headings in this paragraph}

\subsection{Equipment and setup}
\begin{figure}
    %  \centering
    %      \centering
    %      \includegraphics[width=0.98\columnwidth]{figures/setup.jpg}
    %     \caption{Experimental setup}
    %     \label{fig:setup}
    %  \centering
     \begin{subfigure}{0.98\columnwidth}
         \centering
         \includegraphics[width=\textwidth]{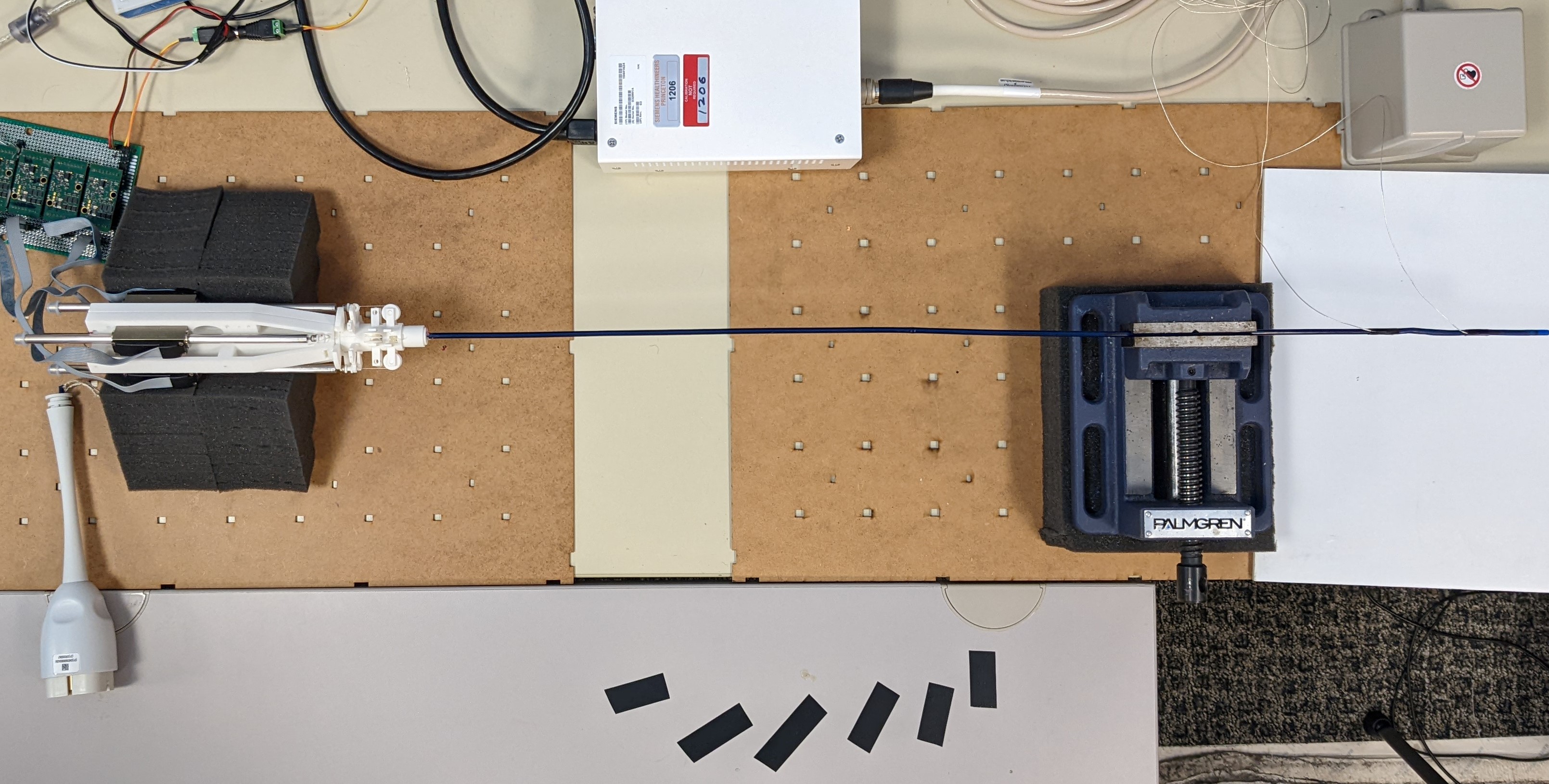}
         \caption{Straight passive proximal section}
         \label{fig:sub-setup1}
     \end{subfigure}
     %\hfill
     \begin{subfigure}{0.98\columnwidth}
         \centering
         \includegraphics[width=\textwidth]{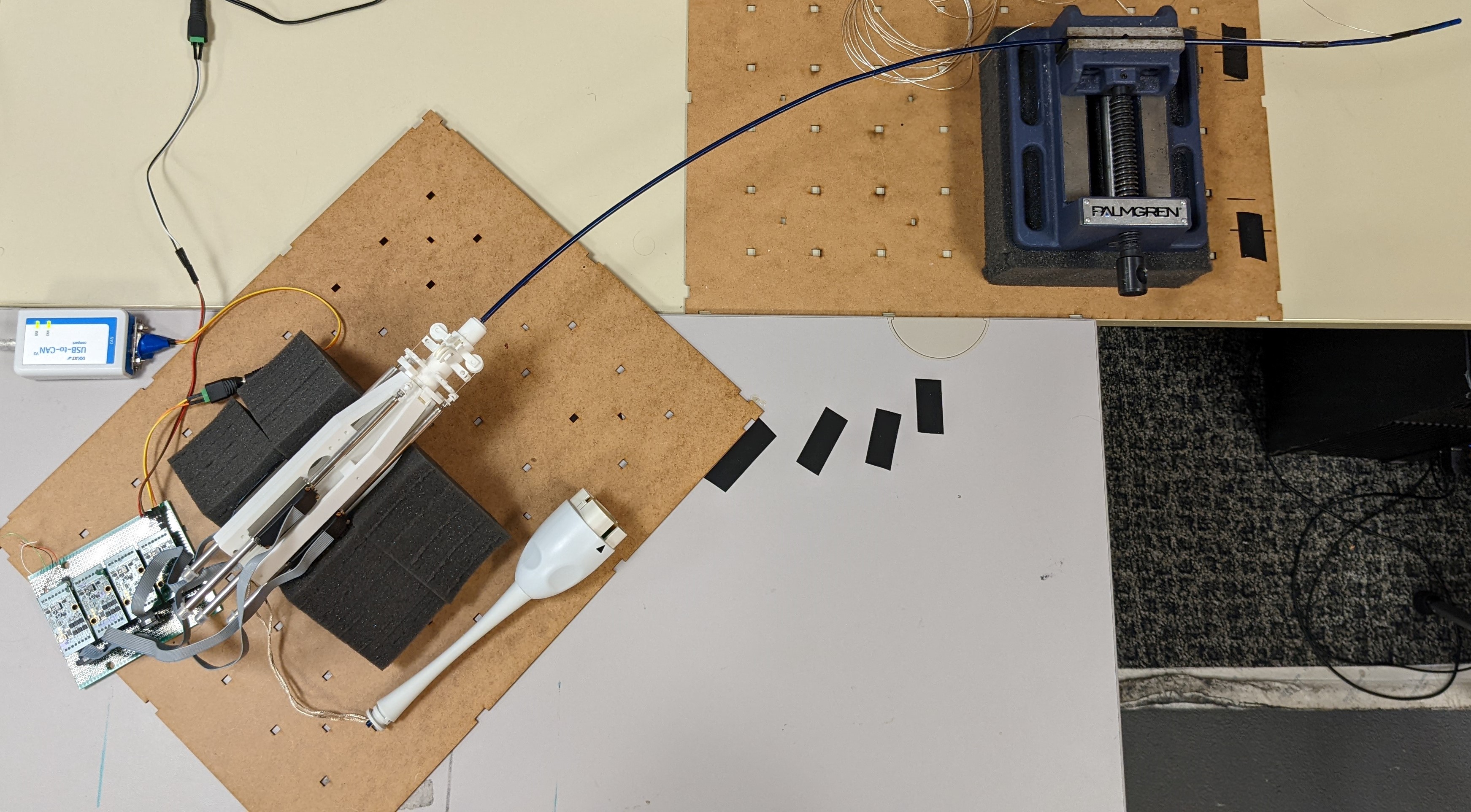}
         \caption{passive proximal section bent to 45$^\circ$ angle}
         \label{fig:sub-setup2}
     \end{subfigure}
        \caption{Experimental setup: testbed of robot has outer cover removed; device is clipped a few centimeters proximal to the articulation section; and black tape marks proximal section angles.}
        \label{fig:setup}
\end{figure}

% 3d tracking how to validate ~~~
% experimental setting
The experimental setup is shown in Fig.\,\ref{fig:setup}. 
The prototype was laid on a table, and the proximal section was lightly clamped just proximal to the bending section using a vise. 
A protractor was used to measure the angles for the catheter body, and tape was used to mark the locations on the table. 
Special care was taken to keep the curve of the proximal section tangent to the axis of the catheter body and of the vise at all angles.
Bending angle was measured using an EM sensor (3D Guidance trakSTAR 2) for validation purposes only, meaning that the EM measurement was not used in the control loop and that the control was performed entirely using the kinematics. 
%\textcolor{red}{What else to mention here? Actuators and motor drivers? Resolution of the EM sensor or of the motor encoders?}

\subsection{Parameter identification}
% \textcolor{blue}{explain that we only use incompressible case but we present both in case someone wants to use this control scheme for a compressible device}
Since compression of our articulation section is negligible, we use the equations and associated parameters for an incompressible device.
All parameters in the equations must be determined empirically. $\textbf{L}_0$, $\textbf{L}_t$, $\textbf{L}_m$, $\textbf{R}$, and $\textbf{K}_t$ can be measured directly. $\mu_1$ and $\mu_2$ are determined heuristically; they are chosen as the smallest constants which result in no slack tendons--as measured from motor current--for a few test inputs. $\textbf{K}$ and $\textbf{D}$ can be approximated by the following parameter identification procedure:
% not sure if its better to detail the procedure here or just say 0.35 mm was found to be a good approximation of the distance d
\begin{enumerate}
  \item Make initial guess for parameters. %\vspace{-8pt}%\yh{use specific terminologies here K, D, etc} \cd{done}
  \item Input motor position trajectories which should achieve desired bending angle as calculated from inverse kinematics and redundant control input. %\vspace{-8pt}  %\yh{Specifically, what input} \cd{done}
  \item Update parameter guesses based on difference between desired and measured bending angle. %\vspace{-8pt} %\yh{how to update? any method name?} \cd{I did not use any special method; I tweaked the number manually. However, any sort of search/optimization minimizing the error would work.}
  \item Repeat steps 2 and 3 until parameter values converge.
\end{enumerate}
%\textcolor{blue}{Should we give the full parameter identification procedure here? If so, we should mention the $\psi_i$ and explain all three major steps of the process. I do not believe we used the $\psi_i$ in the data collection.}
This procedure is also rather heuristic, but it achieved consistent values for $\textbf{K}$ and $\textbf{D}$ with a bisection search. Parameter values are listed in Table\,\ref{Tab:parameters}.
\begin{table}[h!]
\small
\centering
\vspace{0pt}
\caption{Kinematic Parameters}
%\scalebox{0.7}{
\begin{tabular}{||c c c c||} 
%\caption{Mean Absolute Error (MAE) and Standard Deviation (StD) for Tip Position and Bending Angle}
\hline
Parameter & Value & Resolution & Units \\ %[0.5ex] 
\hline\hline
{ $d_1$} & (0.492, -0.0868) & 0.00625 & mm \\
\hline
{ $d_2$} & (0.0868, 0.492) & 0.00625 & mm \\
\hline
{ $d_3$} & (-0.492, 0.0868) & 0.00625 & mm \\
\hline
{ $d_4$} & (-0.0868, -0.492) & 0.00625 & mm \\
\hline
{ $K_b$} & 360.0 & 3.125 & N$\cdot$mm$^2$ \\ 
\hline
{ $k_t$} & 1080 & 5.0 & N \\ 
\hline
{ $L_0$} & 60 & 0.5 & mm \\
\hline
{ $l_{m,1}$} & 20 & 0.5 & mm \\
\hline
{ $l_{m,2}$} & 25 & 0.5 & mm \\
\hline
{ $l_{m,3}$} & 20 & 0.5 & mm \\
\hline
{ $l_{m,4}$} & 30 & 0.5 & mm \\
\hline
{ $l_{t,1}$} & 775 & 0.5 & mm \\
\hline
{ $l_{t,2}$} & 785 & 0.5 & mm \\
\hline
{ $l_{t,3}$} & 775 & 0.5 & mm \\
\hline
{ $l_{t,4}$} & 785 & 0.5 & mm \\
\hline
{ $R$} & 3.0 & 0.1 & unitless \\ 
\hline
{ $\mu_1$} & 0.25 & N/A & unitless  \\ 
\hline
{ $\mu_2$} & 0.25 & N/A & unitless \\ 
\hline
\end{tabular}
%}
 %\vspace*{-10pt}
\label{Tab:parameters}
\end{table}
The tendon locations $d_i$, as seen in Fig.\,\ref{fig:catheter-both}, are given as coordinates $(x,z)$. For simplicity, the tendon locations were assumed to be at 90$^\circ$ increments around the central axis, but this assumption is not necessary to complete the parameter identification. For values which were not measured (i.e., $K_b$ and $d_i$), resolution refers to the smallest step of the bisection used to identify the value.

%%%%%%%%%%%%%%%%%%%%%%%%%%%%%%%%%%%%%%%%%%%%%%%%%%%%%%%%%%%%%%%%%%%%%%%%%%%%%%%%%%%%%%%%%%%%%%%%%%%%%%%%%%%%%%%%%%%%%%%%%%%%%%%%%%
%                                                       EXPERIMENTS                                                              %
%%%%%%%%%%%%%%%%%%%%%%%%%%%%%%%%%%%%%%%%%%%%%%%%%%%%%%%%%%%%%%%%%%%%%%%%%%%%%%%%%%%%%%%%%%%%%%%%%%%%%%%%%%%%%%%%%%%%%%%%%%%%%%%%%%
\subsection{Experiments}

% \textcolor{red}{numbered list of scenarios, what are we doing, what do we expect to see}

The robot was tested under three scenarios of validation which are listed here in brief and explained in detail in the subsections to follow.
\begin{enumerate}
    \item Straight Condition -- sinusoidal input; no passive proximal section angle; baseline and hysteresis compensation
    \item Curved Condition -- sinusoidal input; passive proximal section angles; baseline, re-tension, hysteresis, and re-tension + hysteresis (both) compensation
    \item Dynamic Condition -- re-tension compensation is tested while changing passive proximal section angle
\end{enumerate}

\subsubsection{First Scenario: Straight Condition}

% \textcolor{red}{first scenario: give hamlyn straight condition before this to show mechanism is working in controlled environment and benefits of simple hysteresis compensation in default condition}

% The proposed methods were validated with three trials on a test bed by following a desired sinusoidal output for bending angle in the AP/LR direction. 
The first scenario tests the robot and the redundant controller under the default condition of a straight passive proximal section.
The proposed controller was validated with three trials on a test bed by following a desired sinusoidal input for bending angle in one dimension.
For each trial, the robot followed two cycles of the sinusoid, and the ground truth output angle was measured with an EM sensor, which was only used to evaluate performance.
%the forward kinematics and was not included in the control. 
Tip position and bending angle errors were described as mean absolute error (MAE) and standard deviation (StD).
%both defined as the absolute difference between the quantity calculated from the kinematics and the ground truth, and thus mean absolute error (MAE) and sample standard deviation (StD) are reported. 
Performance was evaluated without and with simple hysteresis compensation. 
The simple hysteresis compensation involved adjusting the input angle by a constant $\pm$ 10 degrees depending on the direction of motion as described in Eq.\,\eqref{eq_hyst}. The backlash width or width of the hysteresis curve ($20^\circ$) was obtained based on the unique physical properties of the catheter\,\citep{lee21hysteresis}.

\subsubsection{Second Scenario: Curved Condition}

% \textcolor{red}{explain that the purpose is to show the decoupling between the input and the error cause: trying to show that error caused by proximal section angle is independent of the input plane}

%experiment itself
We hypothesized that the shape of the passive proximal section affects the accuracy of the kinematics by altering the tensions in the tendons and that any increase in error would be caused by these changes tension. We implemented two types of compensation--one targeted at hysteresis and the other targeted at pre-tension errors caused by the passive proximal section angle--and hypothesized that these are two separate sources of error that would both require compensation.
Data were collected using a test-bed of the prototype under various conditions for two periods of a sinusoidal input with an amplitude of 45$^\circ$ and a frequency of 0.1 Hz. 
Note that the input refers to bending angle, which is then converted to curvature for use in the kinematics.  
For baseline and re-tension compensation, trials were gathered for six passive proximal section angles (15, 30, 45, 60, 75, and 90 degrees) and two bending planes (xy and yz). 
For hysteresis compensation and re-tension + hysteresis compensation, trials were gathered for three passive proximal section angles (30, 60, and 90 degrees) and two bending planes (xy and yz). 
Two bending planes are tested to confirm that error introduced by the passive proximal section angle is independent of the input plane.

\subsubsection{Third Scenario: Dynamic Condition}

% Where the previous scenario examines the effect of the re-tension compensation on the output under static conditions, this scenario examines the re-tension compensation itself under dynamic conditions.
% This scenario examines the re-tension compensation itself under dynamic conditions.
In the previous cases, the re-tension compensation involved giving the robot a few seconds to recover the desired minimum tensions, which was followed by a desired input trajectory; in this case, the controller attempts to maintain the desired minimum tensions for all 30 seconds of each trial, and there is no trajectory input (only the re-tension compensation). The robot is moved manually for approximately 20 seconds to change the passive proximal section angle from 0 to 60 degrees and then allowed to settle for an additional 10 seconds. Data were collected for three trials, both without and with re-tension compensation.  It was expected that a large mismatch between the measured angle and the desired angle of 0 degrees would develop without re-tension compensation.

%%%%%%%%%%%%%%%%%%%%%%%%%%%%%%%%%%%%%%%%%%%%%%%%%%%%%%%%%%%%%%%%%%%%%%%%%%%%%%%%%%%%%%%%%%%%%%%%%%%%%%%%%%%%%%%%%%%%%%%%%%%%%%%%%%%
%                                                   RESULTS                                                                       %
%%%%%%%%%%%%%%%%%%%%%%%%%%%%%%%%%%%%%%%%%%%%%%%%%%%%%%%%%%%%%%%%%%%%%%%%%%%%%%%%%%%%%%%%%%%%%%%%%%%%%%%%%%%%%%%%%%%%%%%%%%%%%%%%%%%
\subsection{Results} \vspace{-0pt}

\subsubsection{First Scenario: Straight Condition Results}

%The hysteresis compensation was to adjust the input angle by a flat amount based on the direction of motion and the width of the uncompensated hysteresis curve. Since the backlash width of our ICE catheter was about 20 degrees, thus we simply propose our compensator as addition of $\pm$ 10 degrees when transition happened.

%The motor positions were controlled with a P controller based on the difference between the redundant control input and the positions measured by the built-in encoders. 

%\yh{We have to reduce this section!: 1)how to test,2) what errors 3) simple hysteresis compensation }

\begin{figure}[t!]
	\centering%\vspace*{-5pt}
	\includegraphics[width = \columnwidth]{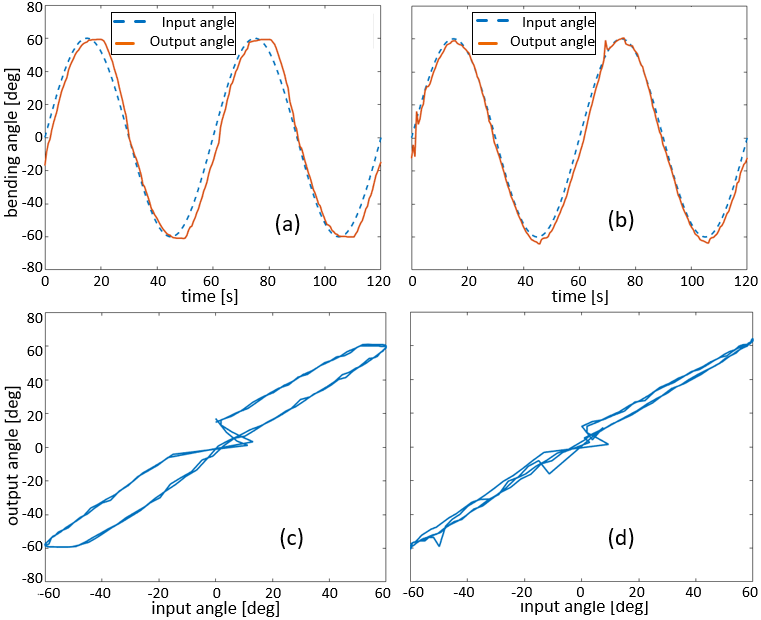}
	\vspace*{-5pt}
	%\caption{An overview of the proposed view-to-view robotic manipulation.\label{fig:workflow}}
	\caption{One result for yz-plane bending: (a) Time vs output angle without compensation (b) Time vs output angle with hysteresis compensation (c)(d) The input angle vs the output angle without/with hysteresis compensation.}
	\label{fig:result}
	\vspace*{-15pt}
\end{figure}

%\cd{moved most of it to previous paragraph, should we define errors here or also previous paragraph?}\yh{RMSE? and STD? What is the error in tables? It should go to above section.}
%\yh{move this to up! Tip position and bending angle errors were both defined as the absolute difference between the quantity  calculated from the kinematics and the quantity measured by the EM sensor.}
%\cd{the error is mean absolute error (MAE) and the standard deviation is sample standard deviation (the one that is usually used)}
Fig.\,\ref{fig:result} shows bending angle without and with the simple hysteresis compensation. 
The control of all tendons simultaneously allowed the removal of tendon slack and thereby deadzone present in conventional catheters. 
The overall performance evaluation is described in Table\,\ref{Tab:error-straight}. The tip pose error (MAE) is reported as $6.1^\circ$. With hysteresis compensation, the position and orientation errors are improved $31\%$ and $47\%$, respectively. 
\begin{table}
\small
\centering
\vspace{0pt}
%\caption{RL-plane error by proximal section angle \& compensation type}
\caption{Bending angle error}
%\begin{tabular}{||p{1.0cm}|p{1.7cm}|p{1.8cm}|p{2.0cm}|p{1.0cm}| p{0.8cm}||}
\begin{tabular}{||c | c | c | c||}
\hline
Angle & Compensation & Error $\pm$ StDev & \% Reduction \\
\hline\hline
 0$^\circ$ & baseline     & 6.11$^\circ$ $\pm$ 4.03$^\circ$      & N/A \\
\cline{2-4} & hysteresis  & 3.26$^\circ$ $\pm$ 2.90$^\circ$      & 46.6 \\
\hline
\end{tabular}
\label{Tab:error-straight}
\end{table}
Bending angle error is reduced by including the simple hysteresis compensation. We also see from Fig.\,\ref{fig:result}c and \ref{fig:result}d that hysteresis is reduced noticeably even with simple compensation.

\subsubsection{Second Scenario: Curved Condition Results}

\begin{figure}[h!]
     \centering
     \begin{subfigure}{0.90\columnwidth}
         \centering
         \includegraphics[width=\textwidth]{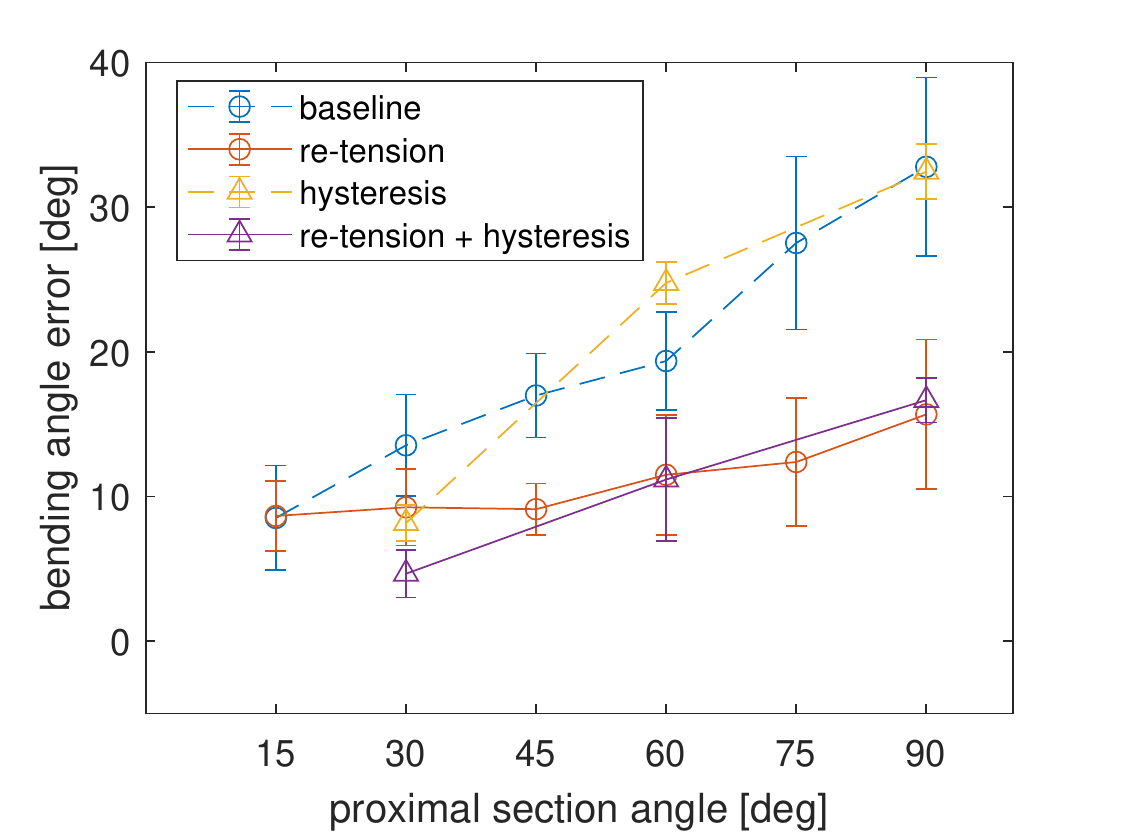}
         \caption{xy-plane bending angle error}
         \label{fig:subfig-R-errorRL}
     \end{subfigure}
     %\hfill
     \begin{subfigure}{0.90\columnwidth}
         \centering
         \includegraphics[width=\textwidth]{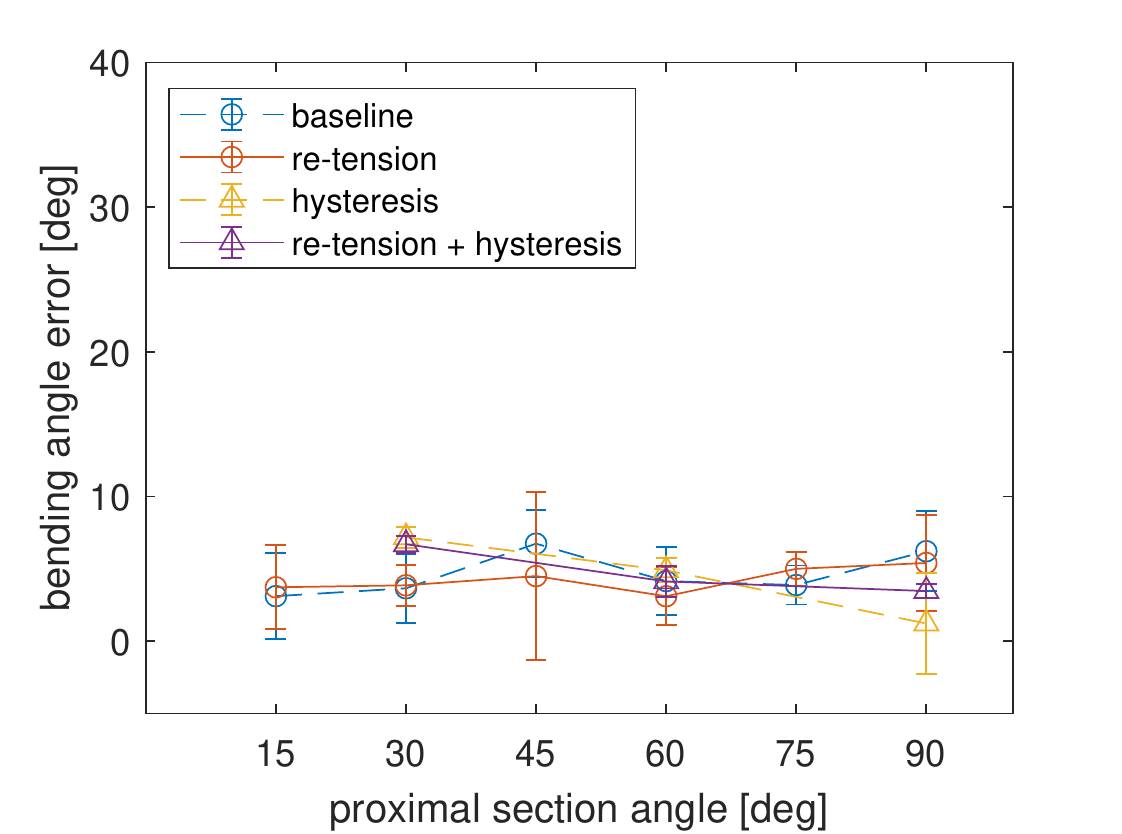}
         \caption{yz-plane bending angle error}
         \label{fig:subfig-R-errorAP}
     \end{subfigure}
        \caption{Bending angle error for different compensation types and proximal section angles (xy-plane input).}
        \label{fig:R-error}
\end{figure}

%This section is divided into subsections based on the bending plane of the input. 
\cd{Until otherwise stated, the following plots and tables for the second scenario involve an input in the xy-plane.} For the xy-plane input, bending occurs in the same plane as the passive proximal section angle, since the passive proximal section angle is created by moving the body to the right. 
%For the AP input, bending occurs perpendicular to the plane in which the passive proximal section is bent. 
Error bars denote the standard deviation over the three trials at that data point.
%\subsection{RL input}
Fig.\,\ref{fig:R-error} shows bending angle error for the six proximal section angles and four compensation types in the xy-plane and yz-plane, respectively.
From Fig.\,\ref{fig:subfig-R-errorRL} it is clear that bending angle error in the xy-plane increases with increasing proximal section angle for all compensation types; however, both compensations with re-tension were less susceptible to this error increase. Error in the yz-plane is unaffected because the passive proximal section is bent in the xy-plane. The error values (RMSE), standard deviations, and percent reduction in error relative to the baseline case are given in Table\,\ref{Tab:error}, and the averages for the proximal section angles with all compensation types are in Table\,\ref{Tab:error-avg}.

\begin{figure}%[!htb]
     \centering
     \begin{subfigure}{0.96\columnwidth}
         \centering
         \includegraphics[width=0.9\textwidth]{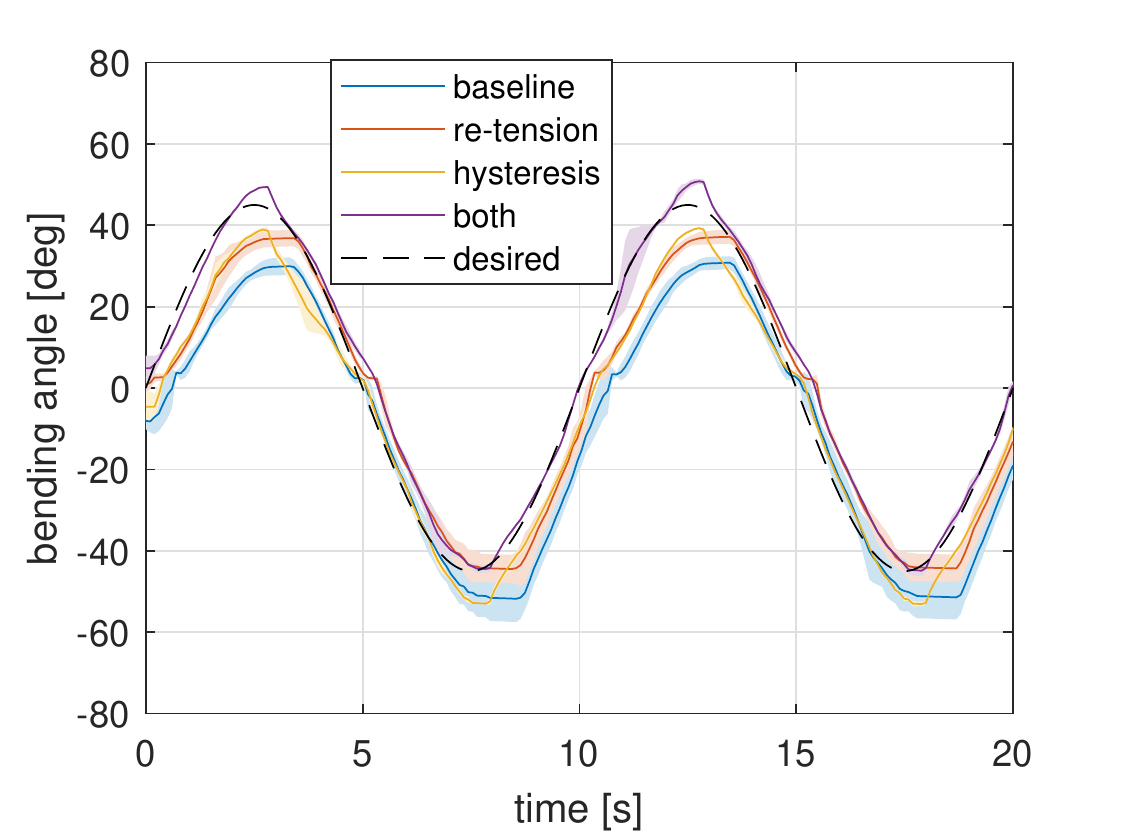}
         \caption{xy-plane angle at 30$^\circ$ proximal section angle}
         \label{fig:subfig-R-bending30RL}
     \end{subfigure}
    %  %\hfill
    %  \begin{subfigure}{0.98\columnwidth}
    %      \centering
    %      \includegraphics[width=\textwidth]{figures/psa-R-RL-angle30AP.eps}
    %      \caption{AP angle at 30$^\circ$ proximal section angle}
    %      \label{fig:subfig-R-bending30AP}
    %  \end{subfigure}
     %\hfill
     \begin{subfigure}{0.96\columnwidth}
         \centering
         \includegraphics[width=0.9\textwidth]{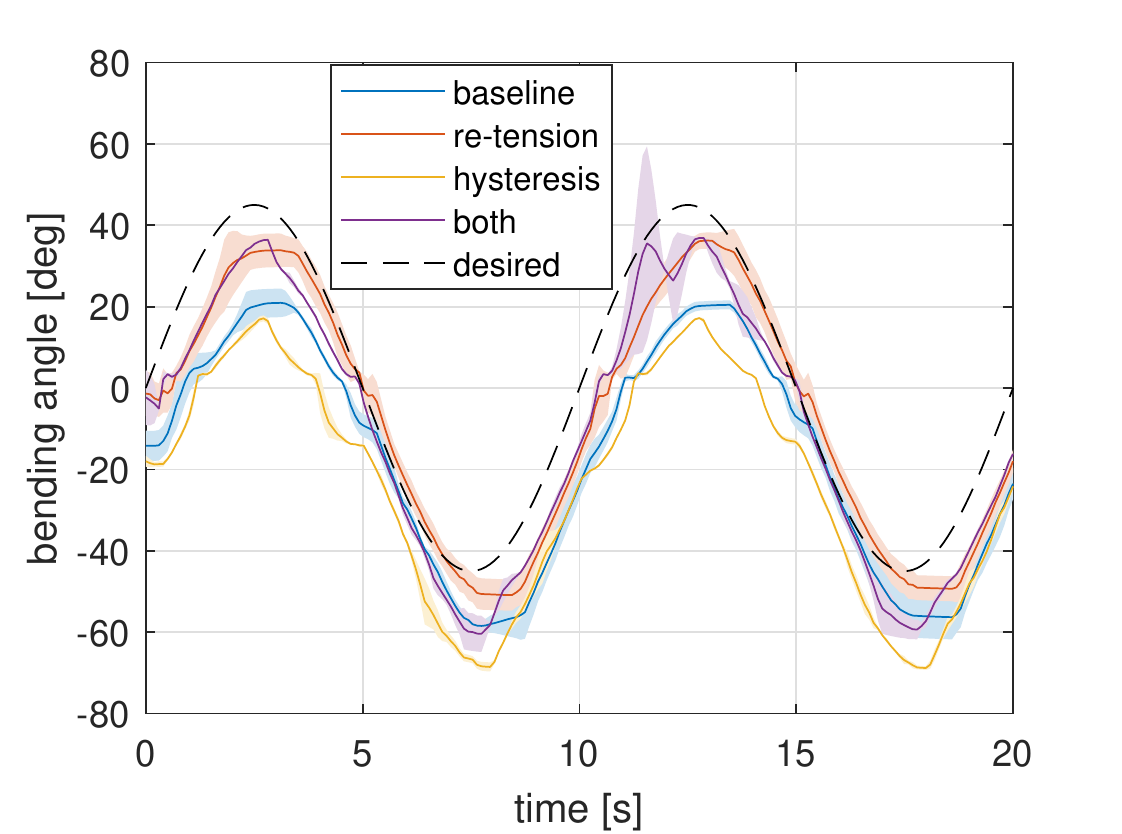}
         \caption{xy-plane angle at 60$^\circ$ proximal section angle}
         \label{fig:subfig-R-bending60RL}
     \end{subfigure}
     %\hfill
    %  \begin{subfigure}{0.98\columnwidth}
    %      \centering
    %      \includegraphics[width=\textwidth]{figures/psa-R-RL-angle60AP.eps}
    %      \caption{AP angle at 60$^\circ$ proximal section angle}
    %      \label{fig:subfig-R-bending60AP}
    %  \end{subfigure}
     %\hfill
     \begin{subfigure}{0.96\columnwidth}
         \centering
         \includegraphics[width=0.9\textwidth]{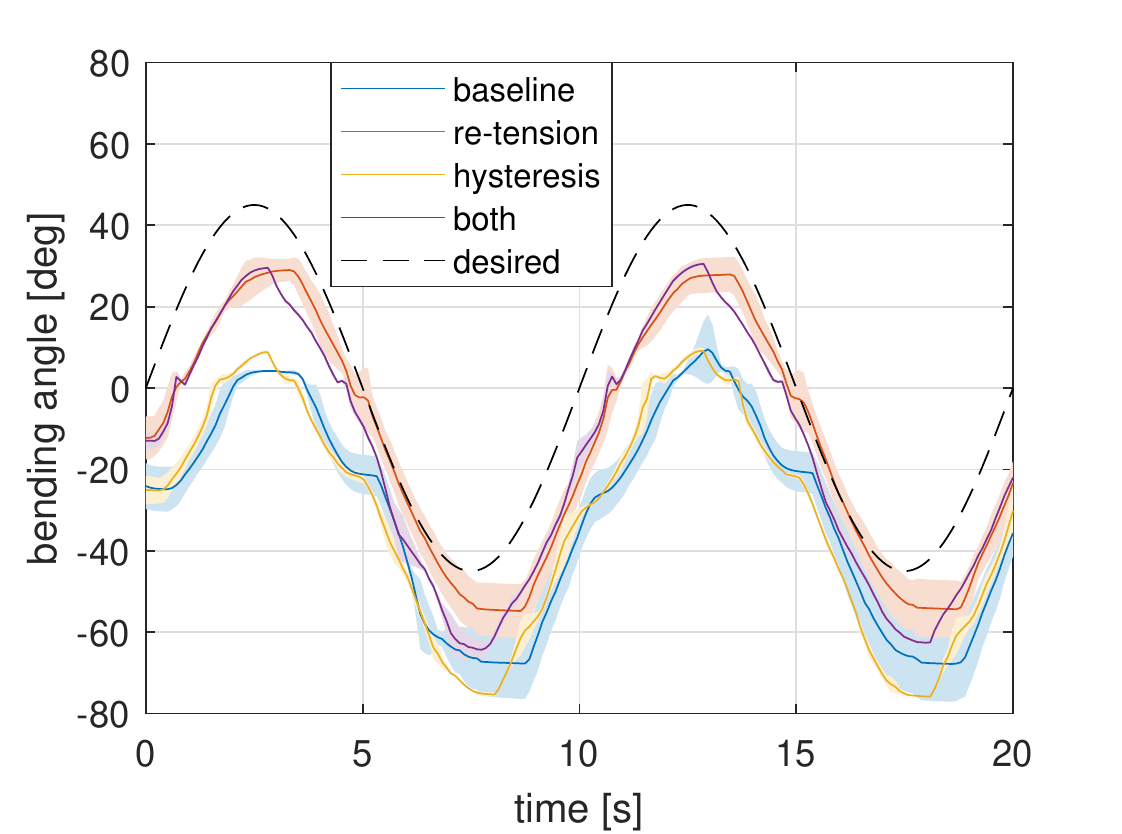}
         \caption{xy-plane angle at 90$^\circ$ proximal section angle}
         \label{fig:subfig-R-bending90RL}
     \end{subfigure}
     %\hfill
    %  \begin{subfigure}{0.98\columnwidth}
    %      \centering
    %      \includegraphics[width=\textwidth]{figures/psa-R-RL-angle90AP.eps}
    %      \caption{AP angle at 90$^\circ$ proximal section angle}
    %      \label{fig:subfig-R-bending90AP}
    %  \end{subfigure}
        \caption{Bending angle for different compensation types at three proximal section angles (xy-plane input).}
        \label{fig:R-bending}
\end{figure}

\begin{table}
\small
\centering
\vspace{0pt}
%\caption{RL-plane error by proximal section angle \& compensation type}
\caption{xy-plane bending angle error (xy-plane input)}
%\begin{tabular}{||p{1.0cm}|p{1.7cm}|p{1.8cm}|p{2.0cm}|p{1.0cm}| p{0.8cm}||}
\begin{tabular}{||c | c | c | c||}
\hline
Angle & Compensation & Error $\pm$ StDev & \% Reduction \\
\hline\hline
 15$^\circ$ & baseline     & 8.52$^\circ$ $\pm$ 3.62$^\circ$      & N/A \\
\cline{2-4} & re-tension   & 8.66$^\circ$ $\pm$ 2.43$^\circ$      & -1.62 \\
\cline{2-4} & hysteresis   & --      & -- \\
\cline{2-4} & both         & --      & -- \\
\hline\hline
 30$^\circ$ & baseline     & 13.54$^\circ$ $\pm$ 3.52$^\circ$      & N/A \\
\cline{2-4} & re-tension   & 9.26$^\circ$ $\pm$ 2.64$^\circ$      & 31.64 \\
\cline{2-4} & hysteresis   & 8.17$^\circ$ $\pm$ 1.23$^\circ$      & 39.66 \\
\cline{2-4} & both         & 4.66$^\circ$ $\pm$ 1.65$^\circ$      & 65.55 \\
\hline\hline
 45$^\circ$ & baseline     & 16.98$^\circ$ $\pm$ 2.90$^\circ$      & N/A \\
\cline{2-4} & re-tension   & 9.12$^\circ$ $\pm$ 1.79$^\circ$      & 46.28 \\
\cline{2-4} & hysteresis   & --      & -- \\
\cline{2-4} & both         & --      & -- \\
\hline\hline
 60$^\circ$ & baseline     & 19.37$^\circ$ $\pm$ 3.38$^\circ$      & N/A \\
\cline{2-4} & re-tension   & 11.50$^\circ$ $\pm$ 4.14$^\circ$      & 40.62 \\
\cline{2-4} & hysteresis   & 24.76$^\circ$ $\pm$ 1.43$^\circ$      & -27.86 \\
\cline{2-4} & both         & 11.17$^\circ$ $\pm$ 4.25$^\circ$      & 42.31 \\
\hline\hline
 75$^\circ$ & baseline     & 27.53$^\circ$ $\pm$ 5.98$^\circ$      & N/A \\
\cline{2-4} & re-tension   & 12.39$^\circ$ $\pm$ 4.45$^\circ$      & 55.00 \\
\cline{2-4} & hysteresis   & --      & -- \\
\cline{2-4} & both         & --      & -- \\
\hline\hline
 90$^\circ$ & baseline     & 32.79$^\circ$ $\pm$ 6.17$^\circ$      & N/A \\
\cline{2-4} & re-tension   & 15.68$^\circ$ $\pm$ 5.18$^\circ$      & 52.19 \\
\cline{2-4} & hysteresis   & 32.45$^\circ$ $\pm$ 1.91$^\circ$      & 1.03 \\
\cline{2-4} & both         & 16.66$^\circ$ $\pm$ 1.54$^\circ$      & 49.19 \\
\hline
\end{tabular}
\label{Tab:error}
\end{table}
% average error over all angles
\begin{table}
\small
\centering
\vspace{0pt}
%\caption{RL-plane error by proximal section angle \& compensation type}
\caption{Average xy-plane bending angle error, average standard deviation, and average \% error reduction for the 30$^\circ$, 60$^\circ$, and 90$^\circ$ trials (xy-plane input).}
%\begin{tabular}{||p{1.0cm}|p{1.7cm}|p{1.8cm}|p{2.0cm}|p{1.0cm}| p{0.8cm}||}
\begin{tabular}{||c | c | c||}
\hline
Compensation & Error $\pm$ StDev & \% Reduction \\
\hline\hline
baseline     & 21.90$^\circ$ $\pm$ 4.36$^\circ$      & N/A \\
\hline
re-tension   & 12.15$^\circ$ $\pm$ 3.99$^\circ$      & 41.48 \\
\hline
hysteresis   & 21.79$^\circ$ $\pm$ 2.48$^\circ$      & 4.28 \\
\hline
both         & 10.83$^\circ$ $\pm$ 1.54$^\circ$      & 52.35 \\
\hline
\end{tabular}
\label{Tab:error-avg}
\end{table}
To give a closer look at how error increases with proximal section angle, bending angle for 30, 60, and 90 degrees is shown in Fig.\,\ref{fig:R-bending}, where shaded regions denote the standard deviation of the trajectory. 
Going from Fig.\,\ref{fig:subfig-R-bending30RL} to \ref{fig:subfig-R-bending60RL} and from Fig.\,\ref{fig:subfig-R-bending60RL} to \ref{fig:subfig-R-bending90RL}, the source of the increasing error becomes more obvious: increases in passive proximal section angle caused an increase in the offset or bias error of the catheter tip. 
This phenomenon was visible in the bending section during experimentation, and can be seen when comparing the tip in Fig.\,\ref{fig:sub-setup1} to the tip in Fig.\,\ref{fig:sub-setup2}. 
Also of note, the quality of the output angle degrades as the offset increases; this can be seen by examining the baseline or hysteresis compensated angle as the offset gets large and may be due to increased friction in the tendon sheath mechanism. 

\begin{figure}%[!htb]
     \centering
    %  \begin{subfigure}{0.96\columnwidth}
         \centering
         \includegraphics[width=0.90\columnwidth]{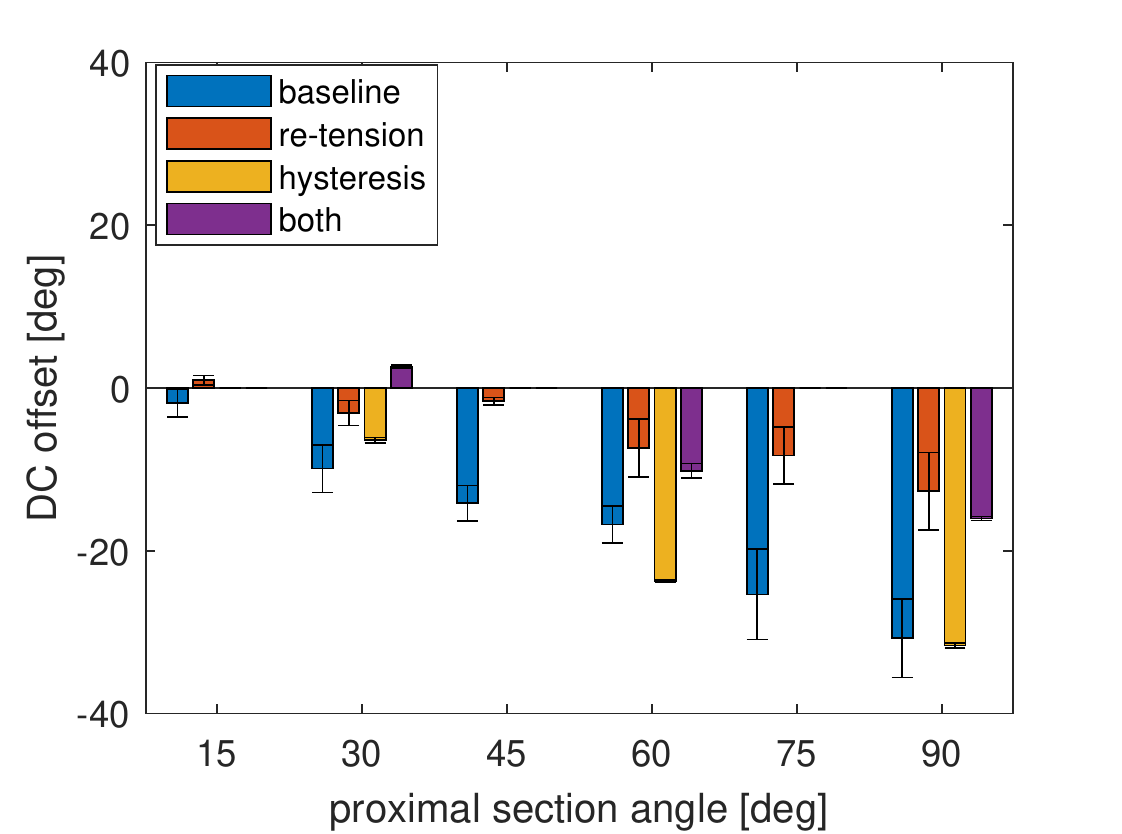}
        %  \caption{RL angle offset for RL bending}
        %  \label{fig:R-offsetRL}
    %  \end{subfigure}
     %\hfill
    %  \begin{subfigure}{0.96\columnwidth}
    %      \centering
    %      \includegraphics[width=\textwidth]{figures/psa-R-RL-offsetAP.eps}
    %      \caption{AP angle offset for RL bending}
    %      \label{fig:R-offsetAP}
    %  \end{subfigure}
        \caption{Bending angle offset for different compensation types and proximal section angles (xy-plane input).}
        \label{fig:R-offset}
\end{figure}

\begin{figure}%[!htb]
     \centering
    %  \begin{subfigure}{0.96\columnwidth}
         \centering
         \includegraphics[width=0.90\columnwidth]{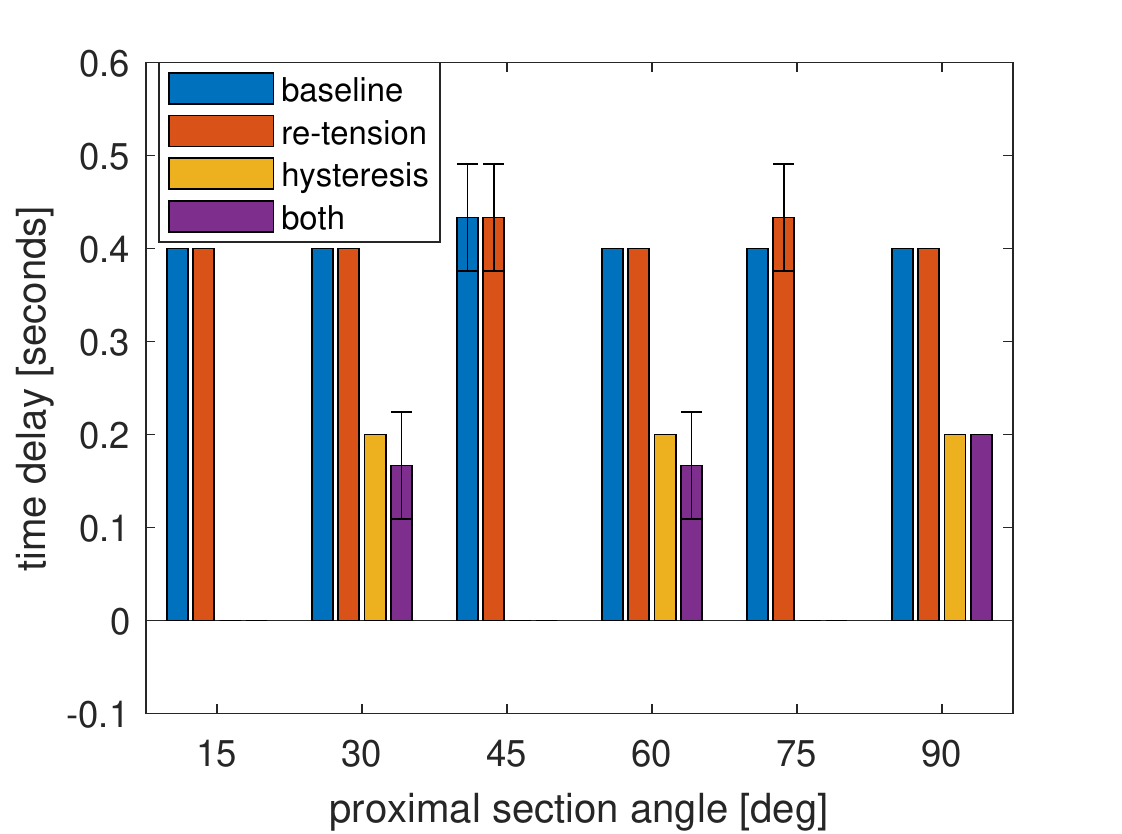}
        %  \includegraphics[width=\textwidth]{figures/psa-R-RL-delayRL.eps}
        %  \caption{RL angle time delay for RL bending}
        %  \label{fig:R-delayRL}
    %  \end{subfigure}
     %\hfill
    %  \begin{subfigure}{0.96\columnwidth}
    %      \centering
    %      \includegraphics[width=\textwidth]{figures/psa-R-RL-delayAP.eps}
    %      \caption{AP angle time delay for RL bending}
    %      \label{fig:R-delayAP}
    %  \end{subfigure}
        \caption{Bending angle time delay for different compensation types and proximal section angles (xy-plane input).}
        \label{fig:R-delay}
\end{figure}

To better show how the re-tension and hysteresis compensation reduce error from different sources, we examine the DC offset and the time delay of the bending angle in Fig.\,\ref{fig:R-offset} and Fig.\,\ref{fig:R-delay}, respectively. Time delay was computed by filtering the output (low pass, cutoff 2 Hz) and finding the cross-correlation between each output and the input. DC offset is the distance of the mean of each filtered output from zero.

The increasing DC offset for increasing proximal section angle in Fig.\,\ref{fig:R-delay} reflects the increasing error seen in Fig.\,\ref{fig:subfig-R-errorRL}. 
% It also mirrors Fig.\,\ref{fig:subfig-R-errorRL} in that the re-tension and re-tension + hysteresis compensation trials are less affected by the offset, just as these compensations both had lower error for increasing proximal section angle. 
It also mirrors Fig.\,\ref{fig:subfig-R-errorRL} in that the re-tension and re-tension + hysteresis compensation trials are less affected by the offset, just as these compensations both had lower error for increasing proximal section angle.
% This suggests that this offset is the primary source of error, and it is caused by the passive proximal section angle.
This suggests that this offset caused by the passive proximal section angle is the primary source of error.
From Fig.\,\ref{fig:R-delay}, it is clear that the simple hysteresis compensation reduces the time delay from hysteresis by half. 
However, Fig.\,\ref{fig:subfig-R-errorRL} shows that the reduction in hysteresis does not reduce the error substantially for increasing proximal section angle, relative to the error reduction from the re-tension compensation. 
This furthers the idea that the DC offset is the larger source of error, especially at higher proximal section angles, and that the effect of the proximal section angle on the bending section should not be ignored.
% \begin{figure}[t!]
% 	\centering%\vspace*{-5pt}
% 	\includegraphics[scale= 0.7]{fig1a.eps}
% 	\vspace*{-0pt}
% 	%\caption{An overview of the proposed view-to-view robotic manipulation.\label{fig:workflow}}
% 	\caption{RL angle error for RL bending for different compensation types}
% 	\label{fig:result1}
% 	\vspace*{-0pt}
% \end{figure}
% \begin{figure}[t!]
% 	\centering%\vspace*{-5pt}
% 	\includegraphics[scale= 0.7]{fig1b.eps}
% 	\vspace*{-0pt}
% 	%\caption{An overview of the proposed view-to-view robotic manipulation.\label{fig:workflow}}
% 	\caption{AP angle error for RL bending for different compensation types}
% 	\label{fig:result2}
% 	\vspace*{-0pt}
% \end{figure}
% \begin{figure}[t!]
% 	\centering%\vspace*{-5pt}
% 	\includegraphics[scale= 0.7]{fig1c.eps}
% 	\vspace*{-0pt}
% 	%\caption{An overview of the proposed view-to-view robotic manipulation.\label{fig:workflow}}
% 	\caption{RL angle error for AP bending for different compensation types}
% 	\label{fig:result3}
% 	\vspace*{-0pt}
% \end{figure}
% \begin{figure}[t!]
% 	\centering%\vspace*{-5pt}
% 	\includegraphics[scale= 0.7]{fig1d.eps}
% 	\vspace*{-0pt}
% 	%\caption{An overview of the proposed view-to-view robotic manipulation.\label{fig:workflow}}
% 	\caption{AP angle error for AP bending for different compensation types}
% 	\label{fig:result4}
% 	\vspace*{-0pt}
% \end{figure}

\begin{figure}%[!ht]
     \centering
     \begin{subfigure}{0.90\columnwidth}
         \centering
         \includegraphics[width=\textwidth]{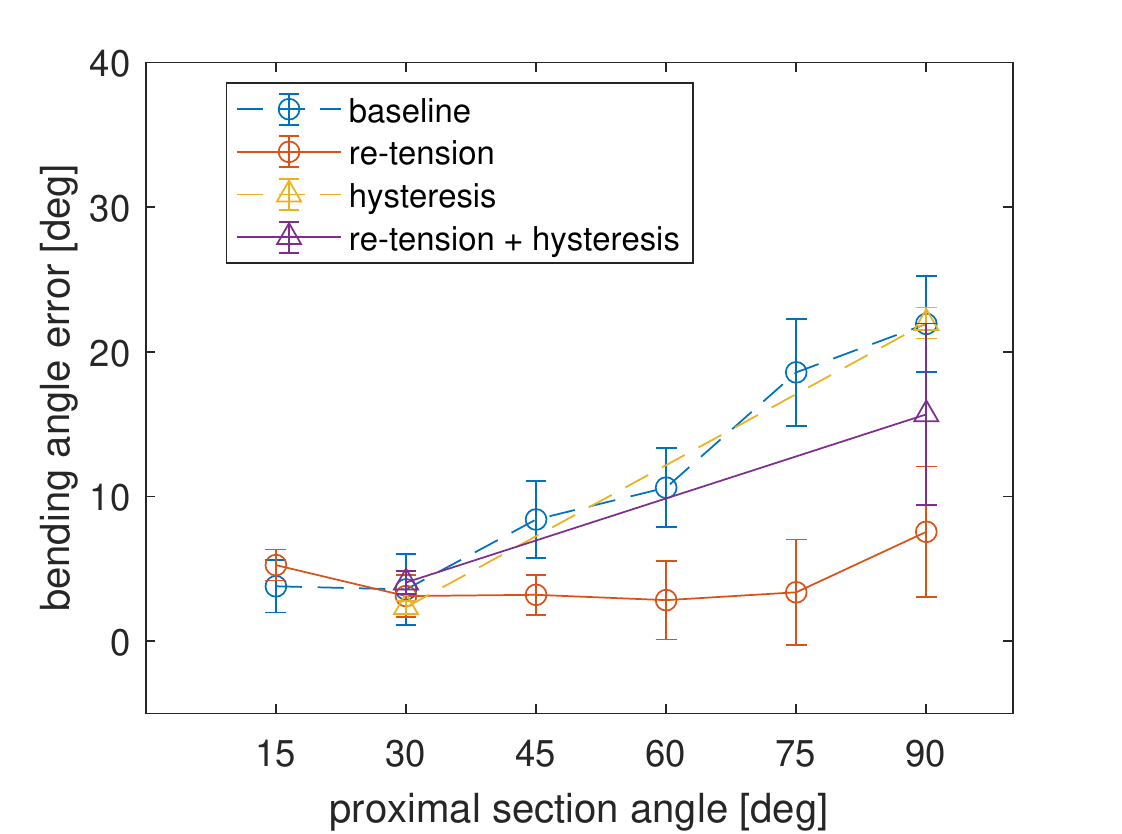}
         \caption{xy-plane bending angle error}
         \label{fig:subfig-R-AP-errorRL}
     \end{subfigure}
     %\hfill
     \begin{subfigure}{0.90\columnwidth}
         \centering
         \includegraphics[width=\textwidth]{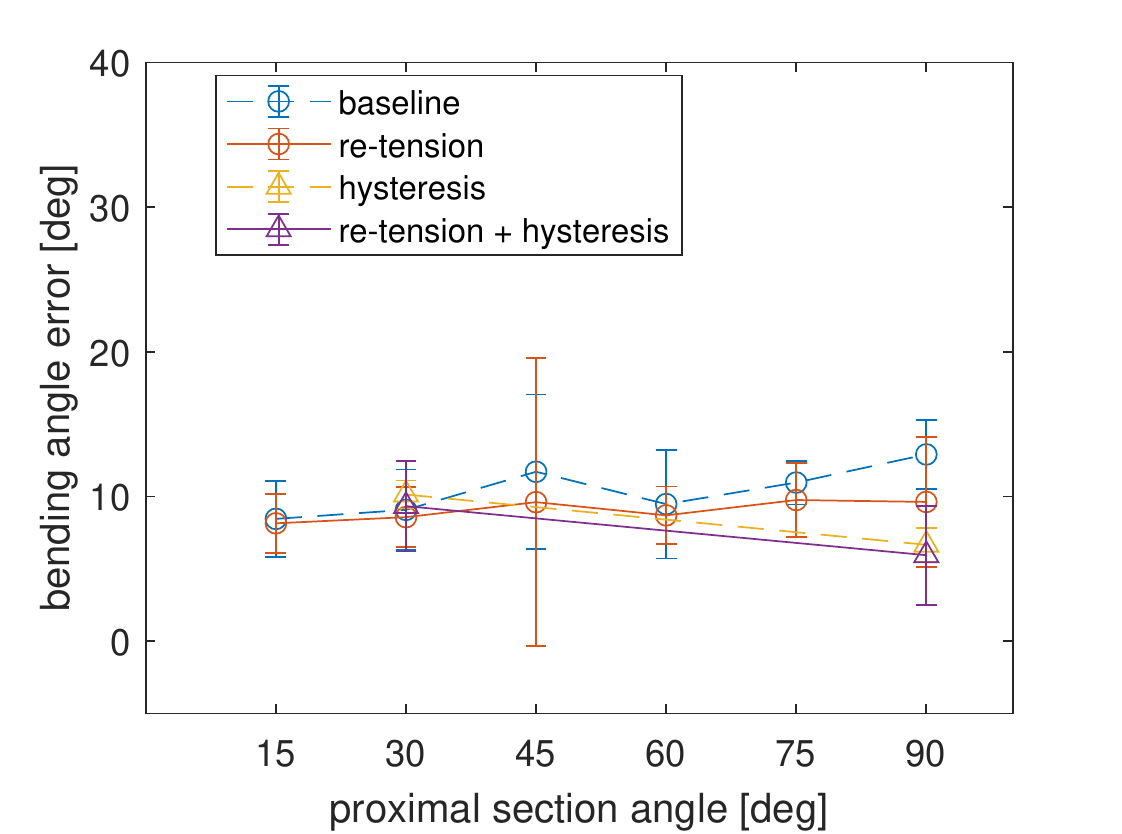}
         \caption{yz-plane bending angle error}
         \label{fig:subfig-R-AP-errorAP}
     \end{subfigure}
        \caption{Bending angle error for different compensation types and proximal section angles (yz-plane input).}
        \label{fig:R-AP-error}
\end{figure}

%\subsection{AP input}
\cd{The remaining figures for scenario two involve an input in the yz-plane.} For the yz-plane input, bending occurs perpendicular to the plane in which the passive proximal section is bent. Fig.\,\ref{fig:R-AP-error} shows that the error in the xy-plane still increases with proximal section angle, even though the input is in the yz-plane. The trials with re-tension compensation still exhibit less error for increasing proximal section angle as well. 
%The trials with hysteresis compensation for 60 degrees were corrupted and thus are not in the following plots. 
\cd{Trials including hysteresis compensation were only collected at the extremes, namely 30 and 90 degrees.}
%\cd{No data was collected for trials with hysteresis compensation at 60 degrees.}
%There is no data for trials with hysteresis compensation at 60 degrees.

\begin{figure}%[!htb]
     \centering
    %  \begin{subfigure}{0.96\columnwidth}
         \centering
         \includegraphics[width=0.90\columnwidth]{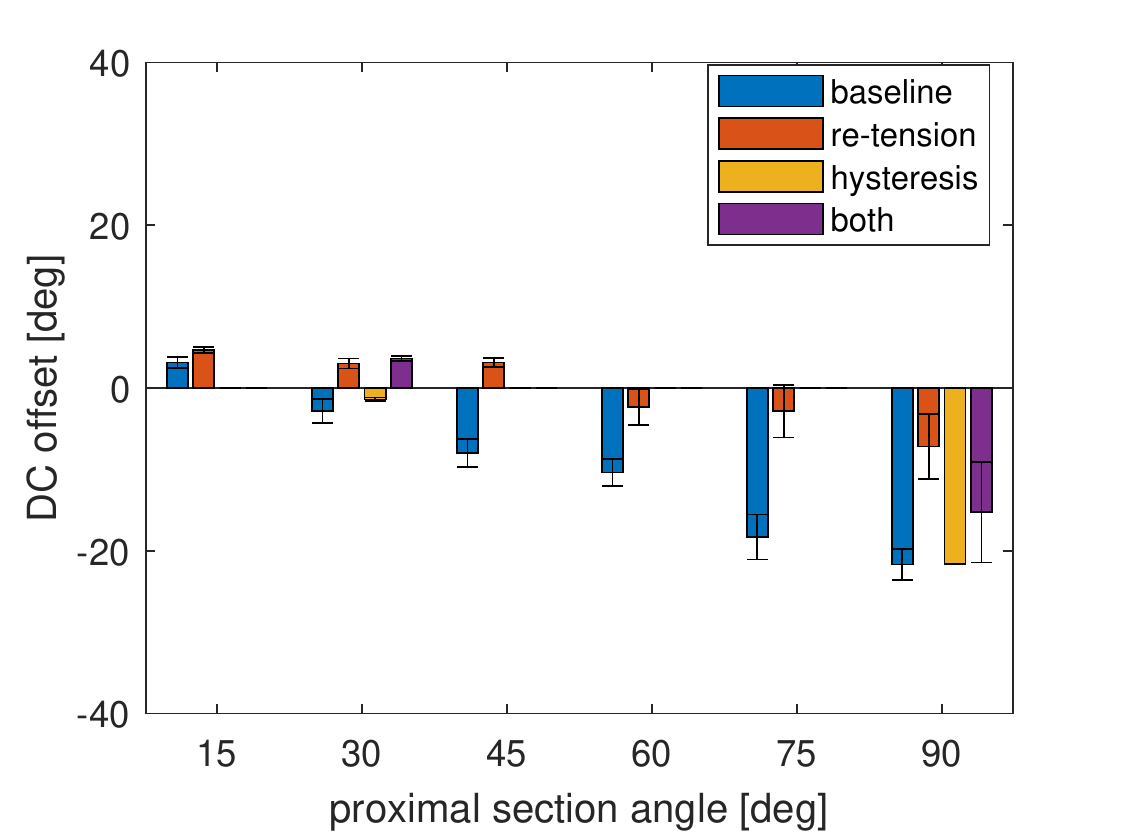}
        %  \includegraphics[width=\textwidth]{figures/psa-R-AP-offsetRL.eps}
        %  \caption{RL angle offset for RL bending}
        %  \label{fig:R-AP-offsetRL}
    %  \end{subfigure}
     %\hfill
    %  \begin{subfigure}{0.96\columnwidth}
    %      \centering
    %      \includegraphics[width=\textwidth]{figures/psa-R-AP-offsetAP.eps}
    %      \caption{AP angle offset for RL bending}
    %      \label{fig:R-AP-offsetAP}
    %  \end{subfigure}
        \caption{Bending angle offset for different compensation types and proximal section angles (yz-plane input).}
        \label{fig:R-AP-offset}
\end{figure}

\begin{figure}%[!htb]
     \centering
    %  \begin{subfigure}{0.96\columnwidth}
    %      \centering
    %      \includegraphics[width=\textwidth]{figures/psa-R-AP-delayRL.eps}
    %      \caption{RL angle time delay for RL bending}
    %      \label{fig:R-AP-delayRL}
    %  \end{subfigure}
     %\hfill
    %  \begin{subfigure}{0.96\columnwidth}
         \centering
         \includegraphics[width=0.90\columnwidth]{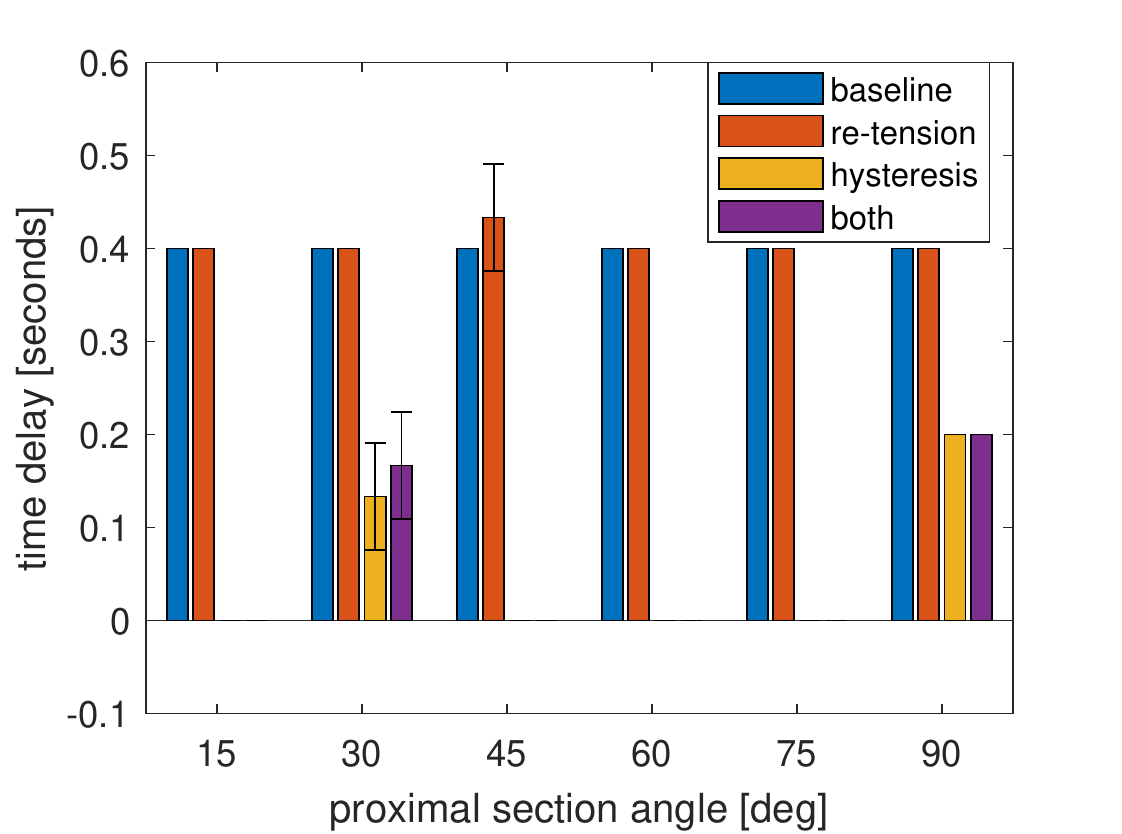}
    %      \includegraphics[width=\textwidth]{figures/psa-R-AP-delayAP.eps}
    %      \caption{AP angle time delay for RL bending}
    %      \label{fig:R-AP-delayAP}
    %  \end{subfigure}
        \caption{Bending angle time delay for different compensation types and proximal section angles (yz-plane input).}
        \label{fig:R-AP-delay}
\end{figure}

We again see the trend on increasing error in the xy-plane reflected in the DC offset of the output in the xy-plane in Fig.\,\ref{fig:R-AP-offset}, although it is less pronounced than in the xy-input case. 
The trials with re-tension compensation have lower offset at 90 degrees compared to those without, but it is difficult to say for certain whether the overall trend from the xy-input case is preserved with the lower offset numbers and missing 60 degree hysteresis data.
Fig.\,\ref{fig:R-AP-delay} shows that the hysteresis compensation reduced the time delay by half as expected from the xy-input case.
% get results for dynamics tests and make plots 

\subsubsection{Third Scenario: Dynamic Condition Results}

\begin{figure}%[!htb]
     \centering
     \begin{subfigure}{0.90\columnwidth}
         \centering
         \includegraphics[width=\textwidth]{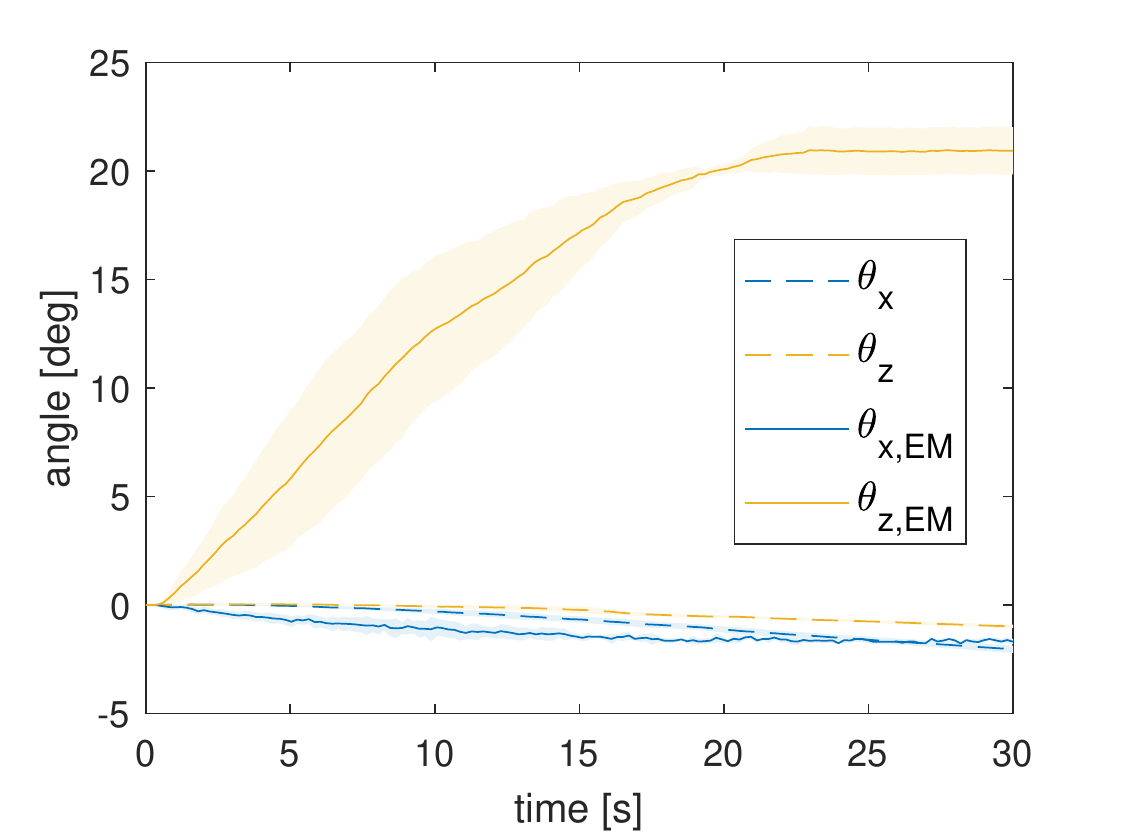}
         \caption{baseline (no re-tension compensation)}
         \label{fig:dynamicOff}
     \end{subfigure}
     \hfill
     \begin{subfigure}{0.90\columnwidth}
         \centering
         \includegraphics[width=\textwidth]{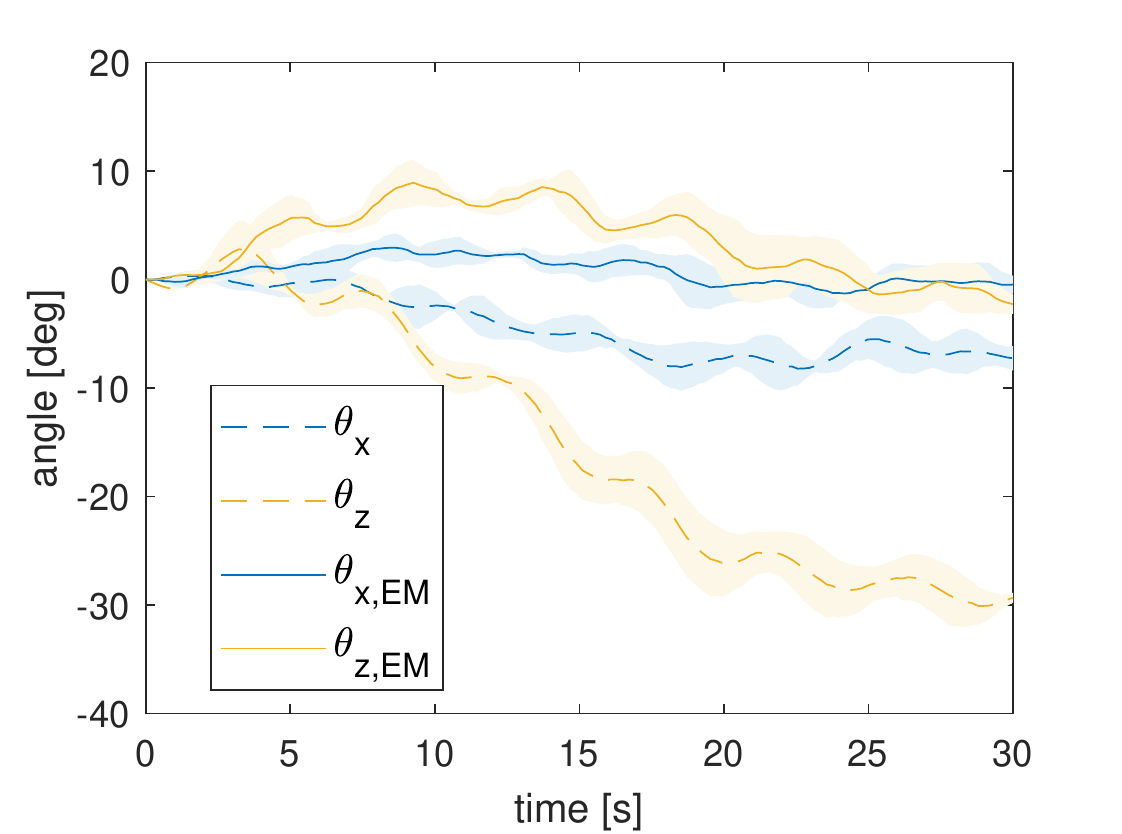}
         \caption{re-tension compensation}
         \label{fig:dynamicOn}
     \end{subfigure}
        \caption{Dynamic tests with re-tension compensation: dashed lines are angles according to the kinematics, and solid lines are measured by the EM sensor.}
        \label{fig:dynamic}
\end{figure}

The effect of introducing the passive proximal section angle in a dynamic scenario is shown in Fig.\,\ref{fig:dynamic}. 
\cd{The bending angle in the xy-plane according to the kinematics and the EM sensor is denoted by $\theta_z$ and $\theta_{z,EM}$ respectively; and the bending angle in the yz-plane according to the kinematics and the EM sensor is denoted by $\theta_x$ and $\theta_{x,EM}$, respectively.}
If the passive proximal section angle had no effect on the kinematics, we would expect to see $\theta_{x,EM}$ and $\theta_{z,EM}$ both remain close to 0 for the duration of the trial in Fig.\,\ref{fig:dynamicOff}. However, in Fig.\,\ref{fig:dynamicOff} we instead see the consequence of increasing the passive proximal section angle from 0 to 60 degrees in the xy-plane without compensation is that the measured angle $\theta_{z,EM}$ accumulates over 20 degrees of offset from 0. 
This agrees closely with the mean error for the baseline trial in the static case in Fig.\,\ref{fig:subfig-R-errorRL}, further supporting that most of the error in the static case comes from this offset error. We also see that this behavior is not captured well by the kinematics, since $\theta_{x}$ and $\theta_{z}$ only deviate by a few degrees. 

% In Fig.\,\ref{fig:dynamicOn}, the re-tension compensation succeeds in bringing $\theta_{x,EM}$ and $\theta_{z,EM}$ back to 0 by the end of the trial. There is a trade-off visible in the trajectory for $\theta_{z,EM}$ in Fig. \ref{fig:dynamicOn}. If the sensitivity of the controller is increased, the trajectory will be held closer to 0, but the oscillations in the trajectory will increase. 
If the control gains are increased, the trajectory will be held closer to 0, but the oscillations will increase.
Table\,\ref{Tab:dynamic-error} gives the measured error in the configuration--in this case deviation from the starting configuration (0,0) by the end of the trial--along with the reduction in error due to the re-tension compensation.
\begin{table}
\small
\centering
\vspace{0pt}
%\caption{RL-plane error by proximal section angle \& compensation type}
\caption{Configuration error after introduction of proximal section angle}
%\begin{tabular}{||p{1.0cm}|p{1.7cm}|p{1.8cm}|p{2.0cm}|p{1.0cm}| p{0.8cm}||}
\begin{tabular}{||c | c | c | c||}
\hline
Angle & Compensation & Error $\pm$ StDev & \% Reduction \\
\hline\hline
$\theta_{x,EM}$ & baseline     & -1.69$^\circ$ $\pm$ 0.12$^\circ$      & N/A \\
\cline{2-4} & re-tension  & -0.47$^\circ$ $\pm$ 0.82$^\circ$      & 72.29 \\
\hline\hline
$\theta_{z,EM}$ & baseline     & 20.94$^\circ$ $\pm$ 1.09$^\circ$      & N/A \\
\cline{2-4} & re-tension  & -2.27$^\circ$ $\pm$ 0.89$^\circ$      & 89.14 \\
\hline
\end{tabular}
\label{Tab:dynamic-error}
\end{table}

%\yh{What is average errors?} \textcolor{blue}{add table}

%%%%%%%%%%%%%%%%%%%%%%%%%%%%%%%%%%%%%%%%%%%%%%%%%%%%%%%%%%%%%%
%                        Discussion                          %
%%%%%%%%%%%%%%%%%%%%%%%%%%%%%%%%%%%%%%%%%%%%%%%%%%%%%%%%%%%%%%
\section{Discussion} \vspace{-0pt}
The design of the robot allowed for the implementation of the redundant controller and additional compensation methods. The current actuators are not strong enough to exceed the breaking force of the magnets, but if more powerful actuators were used, this could become an issue. A good addition to the design to overcome this would be a rail structure which contains the tendon anchors without obstructing the tendons, allowing the motors to reestablish contact with the tendon anchors after a disconnect.

\cd{Although the robot is separable, the control input, compensation methods, and results are applicable to most tendon-driven manipulators with a passive proximal section; and it is unlikely that the separable nature of the device introduced any appreciable error.}
\cd{On a similar note, the robot in this work experienced negligible compression, and thus only the equations for an incompressible bending section were used. That said, the controller does not depend on the assumption of incompressibility, and the relevant equations which were presented can be used for a compressible robot.}
% Thank you for this comment. The compressible case is included for completeness. Although it is not used in our results, it shows that this controller does not depend on the assumption of incompressibility. It is derived so that it can be referred to by the reader if they desire to conduct similar experiments for the compressible case. In other words, compression of our robot was negligible, which is why we only use the incompressible case. However, this work is applicable for other devices where the assumption of incompressibility may not hold.
% \todo{ describe parameter indent shortcomings, and how it is simple because we have few sensors, this is practical estimation and some variability in manufacturing}
\cd{The parameter identification made the practical assumption that the tendons are located at 90 degree increments in order to keep the process simple enough for a small number of sensors.
This simplification is not necessary to use the presented kinematics--the D matrix does not naturally assume 90 degree angles--and relaxing this assumption could yield higher tip position accuracy. 
However, in order to relax this assumption, more sensor information would likely be required during the parameter identification.
% Some effort was even made to identify these parameters originally, but it turns out that they are somewhat diﬀicult to identify reliably without more sensors.
% The positions in D are too small to be measured directly, hence the use of the parameter identification procedure. 
% Relaxing the assumption of orthogonality in our parameter identification resulted in increased error in some cases, most likely because there are more parameters to identify than there are outputs we could observe. 
% Due to this variability, we elected to assume they are orthogonal (as they are supposed to be manufactured) for this study. 
% A thorough examination of the effects of this assumption on tip accuracy for multiple instances of the prototype would be interesting but lengthy, placing it outside the scope of this paper.
}

\cd{A redundant control scheme was used with the constant curvature assumption to resolve and take advantage of this redundancy, in which the control effort is minimized while respecting feasibility constraints (such as requiring a minimum tension for all tendons) as in \citep{platform} but with two key differences. The first difference is that the minimum is easily found analytically since $\mu$ is the only unknown, and so there is no need to optimize in real-time. The second difference is that the axial positions of the tendons do not change appreciably during actuation, meaning that we will not encounter a singular configuration and will not need to reconstruct $\textbf{B}$ as in \citep{platform}.}
% \cd{The final step is to derive the kinematics, but first the case for an incompressible articulation section--which is more relevant in this work--must be derived.}

During the procedure for determining $\mu_1$ and $\mu_2$ in the methods section, we referred to these values as ``pre-tension".
This is because mathematical redundancy in real systems often points to a physical phenomenon, in this case the magnitude of pre-tension or co-contraction of opposing tendons. 
% If we consider Fig.\,\ref{fig:cross} and the incompressible case, the redundancy represents the co-contraction or simultaneous pulling of opposing tendons, such as tendons 1 and 3 or tendons 2 and 4, just as opposing tendons are pulled in muscle contractions. 
If we assume the four tendons are at 90 degree increments and incompressibility, the redundancy represents the co-contraction or simultaneous pulling of opposing tendons, such as tendons 1 and 3, just as opposing tendons are pulled in muscle contractions.
Furthermore, the magnitude of this co-contraction is determined by $\mu_1$ and $\mu_2$. In theory, $\mu_1$ and $\mu_2$ could be any positive value without affecting the configuration of the manipulator. In practice, large values of $\mu_1$ or $\mu_2$ could cause compression of the ``incompressible" bending section, buckling of the passive proximal section, and changes in the stiffness estimates. Thus, it is best to use small values of $\mu_1$ and $\mu_2$ as a method for preventing slack tendons in that any $\mu_1$ or $\mu_2$ value above zero effectively enforces a pre-tension in the corresponding tendons equal to the magnitude of $\mu_1$ or $\mu_2$.
We also suggested keeping the scalable redundant parameters $\mu_1$ and $\mu_2$ constant to avoid affecting other parameter estimates such as bending stiffness. There is room for future experimentation in which these parameters are intentionally varied in order to obtain variable stiffness behavior, akin to impedance control. 

The deflection of the passive proximal section to the right caused an offset of the bending section to the left, irrespective of the plane of the input.
This offset was the primary source of bending angle error and increased with increasing passive proximal section deflection. 
Simple hysteresis compensation alone did not substantially reduce bending angle error, although it did reduce the phase lag of the output. 
Re-tension compensation reduced DC offset and bending angle error, both alone and with hysteresis compensation.  
Compensation methods were kept simple so that the sources of error were not obscured and so that they can be implemented in the absence of a tip sensor; however, there is room for future work involving more complex re-tension or hysteresis compensations, perhaps relying on additional sensors. 

% \yh{Is the re-tension compensation useful in practical settings? Is 5-seconds enough to settle down? What use cases might be useful? For examples, any limitations? Any suggestions for futures?} \cd{I thought future work would go in the conclusion, but it does seem to fit more naturally here? For 5 seconds, it was enough for it to settle, but we do not present the results for different amounts of settling time, so I am not sure we should mention it.}\yh{Future works are related to limitation and some observation during our experiments, so it should be in discussion section. Conclusion section is just wrap-up all contribution etc. I believe 5 seconds settlement time is very conservative action, so we can address, we show the result as a naive way to prove our observation, then say it might be possible to update with better controller such as MPC LQR optimization etc. just examples! You know what i meant?}
The trials with re-tension compensation had reduced error at all angles relative to trials without. By adjusting the tendon tensions to their minimum values, the compensation can be seen as getting much closer to a neutral, unstretched state which the proximal section is assumed to have by kinematic models in the state of the art. 
However, even for trials with re-tension compensation, error still increases for increasing proximal section angle. This is likely because, even with the tendon tensions adjusted, the proximal section is still deformed, deforming the tendon sheaths and the outer casing of the proximal section and articulation section. These deformations are not fully compensated by the re-tension compensation and should be investigated in future works.
% The re-tension compensation was very effective at reducing error from the angle in the passive proximal section; however, it did not remove all of the error introduced. We believe this error comes from deformation of the tendon sheaths and of the plastic outer casing of the proximal section and articulation section, and that this should be investigated in future works.

The amount of time the re-tension compensation was given to reach the desired tension was arbitrary. Since the re-tension occurs before the trial in the second scenario, the time spent re-tensioning does not affect the controller during the trial. This compensation is useful for any application where the proximal shape will remain unchanged for a long period of time, such as after insertion of a heart catheter. For the dynamic case in the third scenario, the re-tension occurs while the proximal section shape is changing. This type of compensation would be useful for applications where the shape is changing over time, such as during insertion of a catheter.

% The amount of hysteresis compensation was based on the width of the hysteresis curve. This curve was relatively constant for different input amplitudes but not exactly constant. There is room for future work in augmenting this simple hysteresis compensation by varying the magnitude slightly based on the input amplitude or some other factor.

The trials with hysteresis compensation tended to have lower error than those without for the lower proximal section angles, and the same or slightly higher error for the higher angles. That hysteresis compensation is less helpful at higher angles is unsurprising, as we would expect errors from the proximal section angle to be the dominant source of error at higher angles. It is a little surprising that hysteresis compensation would cause a slight increase in error at some of the higher angles, and this would suggest that the two sources of error, proximal section angle and hysteresis, are not entirely decoupled. That said, the data presented regarding DC offset and time-shift and the fact the error increase is small both suggest that the phenomena are mostly independent. 
% Also, a less simplistic hysteresis model could alleviate the issue.

% \textcolor{blue}{but its still mostly decoupled, there is room for better representation, but that is a very challenging problem and not the focuss of the work, we asssumed decoupled and it worked well}\yh{Are we supposed to these are decoupled? Why we assume this as decoupled? Let's discuss}
% The re-tension compensation was very effective at reducing error from the angle in the passive proximal section; however, it did not remove all of the error introduced. We believe this error comes from deformation of the tendon sheaths and of the plastic outer casing of the proximal section and articulation section, and that this should be investigated in future works.
The amount of hysteresis compensation was based on the width of the hysteresis curve. This curve was relatively constant for different input amplitudes, but there is room for future work in augmenting this simple hysteresis compensation by varying the magnitude slightly based on the input amplitude or some other factor.

The passive proximal section angle affects the accuracy of the kinematics, and the magnitude of this effect should not be ignored. 
How it is included depends on whether a sensor is present at the tip of the continuum manipulator.
With a tip sensor for closed-loop control, the deviation between the desired output and the measured output can be used to update the kinematics or dynamics to account for the proximal section angle, whether it updates the system parameters directly or simply adds an additional term to the equations. 
Without a tip sensor there is no way to update the kinematics or dynamics real-time without first measuring the behavior for different passive proximal section angles off-line, and so they should be adjusted off-line based on the behavior of the system at different expected configurations.

A strength of this study is that a one-dimensional angle in the passive proximal section causes a clear one-dimensional change in the bending section, but this is also its limitation. 
The effect of multiple bends in different directions, such as two bends in an S-shaped curve, of the passive proximal section should be investigated in future work, though the relationship between the passive proximal section shape and the bending section may not be as simple to characterize.

%%%%%%%%%%%%%%%%%%%%%%%%%%%%%%%%%%%%%%%%%%%%%%%%%%%%%%%%%%%%%%
%                        Conclusion                          %
%%%%%%%%%%%%%%%%%%%%%%%%%%%%%%%%%%%%%%%%%%%%%%%%%%%%%%%%%%%%%%
\section{Conclusion}

We introduced a separable tendon-driven robot manipulator which addresses practical issues in actuation and sterilization. The separable design allows for reuse of all electromechanical components and does not use tendons in the reusable portion, avoiding mechanical wear. The interface between the actuators and the tendons is very transparent and all tendons are actuated, allowing for mitigation of backlash and deadzone. The fan-lock and magnetic interface allow for un-clipping and re-clipping. 

% \textcolor{red}{general paragraph about simplicity of kinematics and redundant control, avoids use of sensors, does not require high fidelity knowledge of parameters a priori which is often unavailable due to variability in manufacturing}
We presented a control scheme which utilizes the simplicity of constant curvature to yield a single solution to the inverse kinematics without the need for real-time optimization. The control scheme can be used without a tip sensor and does not require high fidelity knowledge of system parameters a priori, which is frequently lacking for mass-produced medical devices.

% \yh{How to validate our proposed method? then what is the quantitative numbers for performance?} \cd{You mean we need to bring some info and numbers from the abstract?}\yh{Yes! just rewrite it }
The control scheme was validated along with additional re-tension compensation for proximal section angle and hysteresis compensation.
On average, error was reduced by 41.48\% for re-tension, 4.28\% for hysteresis, and 52.35\% for re-tension + hysteresis compensation relative to the baseline case. 
The re-tension compensation was tested for dynamic changes in the proximal section. The error in the final configuration of the tip was reduced by 89.14\% relative to the baseline case

\section{Disclaimer}\vspace{-0pt}
%\vspace*{-5pt}
%\small
%This feature is based on research, and is not commercially available. Due to regulatory reasons its future availability cannot be guaranteed.
%The concepts and information presented in this paper are based on research results that are not commercially available. 

The concepts and information presented in this paper are based on research results that are not commercially available. Future availability cannot be guaranteed.

\section{Acknowledgement}
Research reported in this publication was supported in part by the National Institute of Biomedical Imaging and Bioengineering of the National Institutes of Health under award number R01EB028278. The content is solely the responsibility of the authors and does not necessarily represent the official 
views of the National Institutes of Health.

%\vspace*{-15pt}
{
	\small
	\bibliography{references_icra22}
}

\end{document}